\documentclass[sigconf]{acmart}

\renewcommand\footnotetextcopyrightpermission[1]{} 
\makeatletter
\renewcommand\@formatdoi[1]{\ignorespaces}
\makeatother

\usepackage{siunitx}
\usepackage{multirow}

\usepackage{amsmath}
\usepackage{balance}
\usepackage{bm}
\usepackage{booktabs}
\usepackage{cleveref}
\usepackage[inline]{enumitem}
\usepackage{eurosym}
\usepackage{mathtools}
\usepackage{units}
\usepackage{tabularx}
\usepackage{placeins}
\usepackage[textsize=tiny]{todonotes}
\usepackage{subcaption}
\usepackage{longtable}
\usepackage{float}
\usepackage{placeins}

\usepackage[utf8]{inputenc}
\usepackage{newunicodechar}

\newunicodechar{￥}{\textyen}
\DeclareTextCommandDefault{\textyen}{%
  \vphantom{Y}%
  {\ooalign{Y\cr\hidewidth\yenbars\hidewidth\cr}}%
}

\newcommand{\yenbars}{%
  \vbox{
     \hrule height.1ex width.4em
     \kern.15ex
     \hrule height.1ex width.4em
     \kern.3ex
  }%
}

\newcommand{\pred}{f_{\operatorname{pred}}}

\newcommand{\valueop}{f_{\operatorname{value}}}
\newcommand{\textcite}[1]{\citeauthor{#1}~\cite{#1}}

\AtBeginDocument{%
  \providecommand\BibTeX{{%
    \normalfont B\kern-0.5em{\scshape i\kern-0.25em b}\kern-0.8em\TeX}}}

\crefname{hypoenum}{question}{questions}
\Crefname{hypoenum}{Question}{Questions}

\begin{document}

\title[]{Towards Explainable Real Estate Valuation via Evolutionary
Algorithms}

\author{Sebastian Angrick}
\author{Ben Bals}
\author{Niko Hastrich}
\author{Maximilian Kleissl}
\author{Jonas Schmidt}
\email{{firstname.lastname}@student.hpi.de}
\affiliation{%
  \institution{Hasso Plattner Institute, University of Potsdam}
  \city{Potsdam}
  \country{Germany}
}

\author{Vanja Dosko\v{c}}
\author{Louise Molitor}
\author{Tobias Friedrich}
\email{{firstname.lastname}@hpi.de}
\affiliation{%
  \institution{Hasso Plattner Institute, University of Potsdam}
  \city{Potsdam}
  \country{Germany}
}

\author{Maximilian Katzmann}
\email{maximilian.katzmann@kit.edu}
\affiliation{%
  \institution{Karlsruhe Institute of Technology}
  \city{Karlsruhe}
  \country{Germany}
}

\begin{abstract}
  Human lives are increasingly influenced by algorithms, which
  therefore need to meet higher standards not only in accuracy but
  also with respect to explainability.  This is especially true for
  high-stakes areas such as real estate valuation.  Unfortunately, the
  methods applied there often exhibit a trade-off between accuracy and
  explainability.

  One explainable approach is \emph{case-based reasoning (CBR)}, where
  each decision is supported by specific previous cases.  However,
  such methods can be wanting in accuracy. The unexplainable machine
  learning approaches are often observed to provide higher accuracy
  but are not scrutable in their decision-making.

  In this paper, we apply \emph{evolutionary algorithms (EAs)} to CBR
  predictors in order to improve their performance.  In particular, we
  deploy EAs to the similarity functions (used in CBR to find
  comparable cases), which are fitted to the data set at hand.  As a
  consequence, we achieve higher accuracy than state-of-the-art deep
  neural networks (DNNs), while keeping interpretability and
  explainability.

  These results stem from our empirical evaluation on a large data set
  of real estate offers where we compare known similarity functions,
  their EA-improved counterparts, and DNNs.  Surprisingly, DNNs are
  only on par with standard CBR techniques.  However, using EA-learned
  similarity functions does yield an improved performance.
\end{abstract}



\maketitle

\renewcommand{\shortauthors}{Angrick et al.}
\pagestyle{plain}

\section{Introduction}
Algorithms have been guiding human decision-making for quite some time, affecting the way we shop or how we select a movie to watch. 
Increasingly, they are employed in sensitive areas such as finance~\cite{dixon2020machine}, medicine~\cite{rajkomar2019machine}, and the legal system~\cite{10.1257/aer.p20161028} as well as real estate valuations~\cite{other_re_ml_with_mape,reviewDNNsOnImmos}.
In such delicate areas algorithms need to fulfill a plethora of requirements: They need to be accurate, trustworthy, and therefore scrutable to the user. Furthermore, they must not exacerbate existing biases or prejudices.
In the context of real estate, these requirements are necessary as setting the price of a house or apartment correctly is of crucial importance to buyer, seller, and the bank providing financing.
However, the algorithms applied there typically fall short of the posed requirements.

On the one hand, \emph{explainable} algorithms outline their reasoning or provide evidence for the decisions they make and are therefore scrutable.
A commonly used explainable approach is built on the following intuition: A human, who is about to make a decision, naturally compares the current situation to similar past experiences and their outcomes.
This process is heavily used by professional real estate appraisers and can be formalized into the \emph{case-based reasoning (CBR)} framework.
There, real estate valuations are estimated as an average of past sales of comparable properties, which then serve as witnesses for the obtained prediction.
This enables humans to check the predictions by looking at the basis on which the algorithm picked the price, allowing them to decide whether the prediction is trustworthy or if it exacerbates certain biases.
In algorithmic applications, the central part of CBR is a \emph{similarity function} that defines how similar the property to be valued is to a property whose value is known.
Similarity functions that have been applied before include simple metrics on attribute vectors~\cite{Baldominos}.
The similarity functions can also be learned from the data set at hand so they are fitted to it~\cite{gayer2007rule}.

On the other hand, \emph{unexplainable} approaches like \emph{deep
  neural networks (DNNs)} have been used to valuate real estate
properties~\cite{reviewDNNsOnImmos}, but they fail the trust
requirement as they only provide an opaque output without any
reasoning.  With the inherent complexity of a DNN, one is not able to
explain in retrospect how a certain output value is produced.  Such
methods are therefore liable to exacerbate existing
biases~\cite{o2016weapons,racist-machine, noor2020can,
  lazovich2020does}, which can only be discovered by statistical
analysis of many predictions.  A user cannot check an individual
prediction for them.  Nevertheless, DNNs have several advantages.
Standard architectures can be applied to a variety of areas without
relying on domain experts, which can save cost and time resources, and
at the same time they are observed to provide high
accuracy~\cite{reviewDNNsOnImmos}.

In this paper, we improve upon previous explainable methods by using \emph{evolutionary algorithms (EAs)}, which are getting more and more popular in the field of artificial intelligence~\cite{som-e-17}.
In particular, for CBR predictors we apply EAs to find good similarity
functions that are fitted to the considered data set, allowing us to
improve the accuracy of CBR while maintaining its explainability.
Moreover, while previous approaches based on similarity functions do
not scale to large data sets as they rely on maximum likelihood
estimation and numerical methods~\cite{MillarRussellB2011MLEa}, our
approach reduces the complexity for a single prediction from linear to
logarithmic.  This improvement can be attributed to the use of
geometric data structures that allows us to find similar objects
efficiently.

To evaluate the resulting adaptation, we perform an empirical
evaluation on a large data set of Japanese real estate offers
\cite{lifull-dataset}, where we compare the different approaches.  The
results show that, surprisingly, the unexplainable DNNs do not beat
known CBR approaches.  However, our EA-assisted CBR method outperforms
all others, yielding an explainable approach with higher accuracy.

In our analysis, we further identify an advantage that CBR has over
DNNs.  One of the fundamental rules in real estate valuation is that
the value of a property is heavily influenced by its location.  Hence,
its surrounding properties are an important indicator for its price.
While DNNs are a general framework that can be used to achieve good
results without relying on domain knowledge, they seem to be unable to
make these local relationships, as this information is only implicitly
present in the weights learned during training. On the other hand, CBR
makes explicit connections between properties, which turns out to
intrinsically capture this important aspect.  When providing DNNs with
information about surrounding properties, they become much more
accurate.

\paragraph{Structure of the Paper} This work is structured as follows. In \Cref{sec:preliminaries}, we introduce preliminary notation and used metrics. We discuss various methods to evaluate real estate properties in \Cref{sec:real-estate-valuation}. In particular, this includes the CBR and DNN approach, as well as our proposed method to utilize EAs to improve the underlying similarity functions. We evaluate our work in \Cref{sec:results} and conclude it in \Cref{sec:conclusion}.

\subsection{Related Work}
\label{sec:related-work}

Early work in the computerized assessment of residential properties was mainly conducted using linear regression techniques, e.g. \cite{carbone1977feedback}.
They were superseded by \emph{hedonic price models}, which determine the price by estimating the effect of each characteristic a property has.
These well-researched models have remained the dominant technique for decades~\cite{glumac2018real}.
However, since hedonic models assume that a property's price is the sum of its desirable attributes, they are limited.
With the advent of more general machine learning methods like \emph{Support Vector Regression}, \emph{Random Forests}, and \emph{Deep Neural Networks}, these have also been applied to real estate valuations.
While these methods are more accurate and outperform the classical hedonic models~\cite{other_re_ml_with_mape, Valier2020, seya2019comparison, afonso2019housing, adgeo-45-377-2018, roy}, they lack explainability due to their black box character.

Moreover, it has been observed that the explainable approach of taking the average of the $k$-nearest neighbors beats Support Vector Regression and \emph{Multi-layer Perceptrons} (simple neural networks), while \emph{Regression Trees} perform even better~\cite{Baldominos}.

Further explainable approaches were obtained by introducing the concept of \emph{case-based reasoning} and \emph{rule-based reasoning} (a hedonic regression model) to real estate valuations~\cite{gayer2007rule}. 
There, it was observed that the CBR approach works better on a database of rental prices, while the rule-based approach was superior on a sales database.  
We extend their work by learning a more complex similarity function for case-based reasoning using EAs, which are applicable to large data sets.

EAs have been previously applied to real estate valuation~\cite{real-estate-ga}, where they are utilized to learn a hedonic model. 
To the best of our knowledge, EAs have not been used to learn similarity functions for real estate valuation before.

We work with a data set of real estate offers from the Japanese "LIFULL HOME'S Data Set"~\cite{lifull-dataset}, which has been used before to compare DNNs with \emph{Kriging}, a generalization of Gaussian modeling to spatial data~\cite{seya2019comparison}.
In particular, they look at the \emph{nearest neighbor Gaussian processes model}, which enables the application of Kriging to big data sets.
In contrast to our contribution, they examine rent price predictions on data set samples with sizes from \(10^{4}\) to \(10^{6}\).
They find that for the largest sample size \(10^{6}\) Kriging and DNNs perform similarly but DNNs are superior on the extreme ends of the price range.
While they only consider simple architectures, like previously used neural networks with up to 5 layers~\cite{raju2011development}, we examine more complex architectures. 
In particular, we use the state-of-the-art TabNet~\cite{tabnet}, a high-performance and interpretable canonical deep tabular data learning architecture.
In the context of real estate, two further architectures are proposed on
Kaggle\footnote{\url{www.kaggle.com}}, a data science community offering machine
learning competitions: A DNN published in a competition on real estate
prices~\cite{house_prices_kaggle} and one published in another real
estate price prediction challenge with the intent to serve as a baseline for
DNNs with tabular data~\cite{luckeciano_kaggle}. Throughout this work, we denote the former architecture by \emph{Kaggle Housing} and the latter by \emph{Kaggle Baseline}, both of which are also considered in our evaluation.

\section{Preliminaries}
\label{sec:preliminaries}
Let \(\mathbb{N}\) and $\mathbb{R}$ denote the set of natural and real numbers, respectively.
For \(n \in \mathbb{N}\), we define $[n] = [1,n] \cap \mathbb{N}$.
Let \(\overline{\mathbb{R}} = \mathbb{R} \cup \{-\infty, \infty\}\) be the extended real numbers.
For a vector \(\bm{v}\) we denote by \(v_i\) the \(i\)-th entry of \(\bm{v}\).

\subsection{Error Measures}
\label{sec:error-measures}

In order to evaluate the quality of a predictor, we take a set of~$n$ objects, whose values $\bm{y} \in \mathbb{R}^n$ are known.  We
then give the objects (but not the values) to the predictor, which
then generates a set of predictions $\bm{\hat{y}} \in \mathbb{R}^n$.
Now we want to measure how well the predictions match the actual
values of the objects.  One such measure is given by the \emph{mean
  percentage error (MPE)}, which is defined as
\begin{align*}
  \text{MPE}(\bm{y}, \bm{\hat{y}}) = \frac{\sum_{i \in [n]} 1 - \hat{y}_i/y_i}{n}.
\end{align*}
The MPE can be used to determine whether the predictor has a bias,
i.e., tends to produce predictions that are larger or smaller than the
ground truth.  One disadvantage, however, is that prediction errors
may cancel each other.  This can be avoided by using the well-known
\emph{mean absolute percentage error (MAPE)}, that is,
\[
  \text{MAPE}(\bm{y}, \bm{\hat{y}}) = \frac{\sum_{i \in [n]} |1 - \hat{y}_i / y_i|}{n}.
\]
Both measures are intuitively understandable, and in contrast to other
loss functions allow us to compare the success on different price
ranges, since the errors are given in percentage and are therefore not
relative to the size of the numbers in the data itself.  In
particular, the MAPE is standard in machine learning and was used in
the context of predicting real estate prices
before~\cite{other_re_ml_with_mape, seya2019comparison}.

\section{Real Estate Valuation}
\label{sec:real-estate-valuation}
In this section, we introduce a well-known prediction scheme that is already being applied to real estate valuation~\cite{gayer2007rule, Baldominos}. We propose extensions to make the method work on large data sets, and explain how EAs can be used to improve its performance.  Moreover, we briefly explain how DNNs are used to predict real estate prices~\cite{reviewDNNsOnImmos, seya2019comparison, afonso2019housing}.

\subsection{Valuation with Case-Based Reasoning}
\label{sec:case-based}

\subsubsection{Case-Based Reasoning}

When applying CBR, we gain knowledge about a new property by considering similar properties whose valuations are known.
Formally, this means that we have a set of properties $P$, divided into two sets $P_u$ and $P_v$, denoting \emph{unvalued} and \emph{valued} properties, respectively. The valuations of properties in $P_v$ are given by a function $\valueop \colon P_v \to \mathbb{R}$, while a symmetric similarity function \(s\colon P \times P \to \overline{\mathbb{R}}_{\ge 0}\) indicates how similar two properties in $P$ are.
Now, we predict the value of an unvalued property \(p_u \in P_u\) as
\begin{align}
  \label{eq:wap}
\pred(p_u) = \frac{\sum_{p_v \in P_v} s(p_u,p_v) \cdot \valueop (p_v)}{\sum_{p_v \in P_v} s(p_u, p_v)}.
\end{align}
This definition is well known and often referred to as \emph{weighted average prediction} \cite{rest.88.3.443}.

The similarity function $s$ can be used to represent a natural notion of closeness.
In the real estate case an intuitive choice is the squared inverse geographical distance, which we denote as \emph{location-based similarity (LBS)}.
Another commonly used approach is to develop the similarity function manually based on the attributes of the properties.
One example is the unweighted Euclidean distance between attribute vectors~\cite{Baldominos}.
This function can be generalized by assigning weights to the attributes and allowing the use of different metrics.
The resulting parameters can then be adjusted to fit a given data set using machine learning techniques~\cite{gayer2007rule}.
However, this method is not computationally feasible for large data sets, since the expensive objective function has to be computed often~\cite{MillarRussellB2011MLEa}.

\subsubsection{Extending the CBR Approach}

We propose finding similarity functions via EAs.
This offers a number of advantages:
\begin{enumerate*}[label=(\arabic*)]
\item the similarity function is still fitted to the data set at hand and therefore promises better performance,
\item EAs are very flexible and do not rely on special properties of the function to learn, such as differentiability, and
\item since EAs need to evaluate the objective function less often than comparable optimization techniques, they are better suited to large data sets.
\end{enumerate*}
We refer to this method as the \emph{CBR+EA} approach.

Formally, we encode a property $p \in P$ as an \emph{attribute vector} \(a(p) \in \mathbb{R}^n\), where, for $i \in [n]$, the entry $a_i(p)$  corresponds to the $i$-th attribute of the property.
We propose to learn the inverse of a weighted quasi-norm.  That is, we learn \(q \in \mathbb{R}^+\) and a weight vector \(\mathbf{w} \in \mathbb{R}^n\) and then define the similarity for two properties $p_1, p_2 \in P$ to be
\[
  s^{q, \bm{w}}(p_1, p_2) = {\bigg(\sum_{i \in [n]} w_i {(|a_i(p_1) - a_i(p_2)|)}^q \bigg)}^{-1/q}.
\]
This measure has four key advantages:
\begin{enumerate*}[label=(\arabic*)]
  \item it has previously demonstrated good performance as a distance measure in high dimensional spaces~\cite{aggarwal2001surprising},
  \item it can represent a wide variety of similarity functions due to weighting and a variable exponent,
  \item it only exposes a reasonable number of parameters (namely the number of attributes plus one), and,
  \item by introducing \(q\), we extend the function learned by \cite{gayer2007rule} while by allowing weighting, we extend the Euclidean norm employed by \cite{Baldominos}.
\end{enumerate*}

\subsubsection{Application to Large Data Sets}
\label{sec:pre-post-filter}
Note that, by Equation~\eqref{eq:wap}, determining a price for a single unvalued property $p_u$ involves computing the similarity function for all objects in the data set.
This may become infeasible when using large data sets.
Moreover, even if all computations are performed, the resulting amount of data may be too large for a human to grasp in order to verify the decision of the algorithm.
To mitigate both problems, we extend upon our model by introducing \emph{filtering} as well as \emph{pre-} and \emph{post-selection}.

In filtering, we learn a vector \(\bm{f} \in \overline{\mathbb{R}}_{\ge 0}^n\) that we can wrap around a given similarity function \(s\) such that we get a filtered similarity function \(s_{\bm{f}}\).
For two properties $p_1, p_2 \in P$ we define
\begin{align*}
  s_{\bm{f}}(p_1, p_2) =
  \begin{cases}
    0, &\text{if $\exists i \in [n] \colon |a_i(p_1)-a_i(p_2)| \ge f_i$ and} \\
    s(p_1, p_2), &\text{otherwise}.
  \end{cases}
\end{align*}
Intuitively, the filtering ensures that two properties are considered not similar at all when their attribute vectors differ by too much in at least one component.

Our second extension of the CBR approach is motivated by the fact that the value of a property $p_u \in P_u$ is mostly influenced by its location~\cite{Kryvobokov2007}, making it unlikely that properties with a large geographical distance to $p_u$ contribute reasonable values to the prediction.
Therefore, we introduce \emph{pre-selection}, which works as follows.
When predicting the value of $p_u$, we want to compute the function in Equation~\eqref{eq:wap}, which involves computing the similarity between $p_u$ and \emph{all} valued properties $p_v \in P_v$.  To avoid this, we approximate the weighted average prediction $\pred$ by only considering the subset of properties in $P_v$ containing the $k \in \mathbb{N}_{>0}$ properties that are geographically closest to $p_u$ (among those whose geographical distance to $p_u$ is below a certain threshold $r \in \mathbb{R}_{\ge 0}$).
Pre-selection can be implemented efficiently using geometric data structures like R*-trees~\cite{rstar}.
This reduces the number of required computations of the similarity function, which in turn further reduces the computational cost of the fitness function.

Finally, in order to reduce the number of objects that influence the price, therefore improving human verifiability, we introduce \emph{post-selection}.
There, the weighted average (\Cref{eq:wap}) is only taken over the \(m \in \mathbb{N}_{>0}\) valued properties (which also pass pre-selection) that are most similar to~\(p_u\).

In total, the resulting prediction function is given by
\begin{align*}
  \pred^{q, \bm{w}, \bm{f}, m, k, r}(p_u) = \frac{\sum_{p_v \in P_v'} s_{\bm{f}}^{q, \bm{w}}(p_u, p_v) \cdot \valueop (p_v)}{\sum_{p_v \in P_v'} s_{\bm{f}}^{q, \bm{w}}(p_u, p_v)},
\end{align*}
where $P_v'$ is the subset of properties of $P_v$, that fulfill the requirements of pre- and post-selection with parameters $m, k$, and $r$, and $s_{\bm{f}}^{q, \bm{w}}$ is the above mention weighted $s^{q, \bm{w}}$ similarity function after applying the filter $\bm{f}$.

\subsection{Searching Similarity Functions with EAs}

We now use an EA to find good values for the parameters of this similarity function.
In particular, we consider a tuple $(q, \bm{w}, \bm{f}, m, k, r)$ of the parameters as an \emph{individual}.
Throughout several \emph{generations}, a \emph{population} of multiple individuals is altered using \emph{mutation} and \emph{crossover}.
All parameters are mutated separately with a fixed probability; if selected for mutation, the new value is sampled from a normal distribution centered at the current value.
We employ \emph{uniform crossover}, which means that each parameter is taken from either parent with the same probability.
For parameters that are vectors ($\bm{w}$ and $\bm{f}$), the crossover is applied element-wise.
The \emph{fitness} of an individual, i.e., the performance of the prediction function obtained when using the corresponding parameters, is evaluated as follows.  The given data set is split into two sets containing valued and unvalued properties, respectively.  We then predict the values of all unvalued properties by using the similarity function defined by the parameters of the individual on the valued properties.  The MAPE (see Section~\ref{sec:error-measures}) of these predictions then represents the fitness value.
Among the individuals encountered throughout all generations, the one with the highest fitness value is then used to obtain the final similarity function.

\subsection{Valuation with Deep Neural Networks}
\label{sec:neural-networks}

Deep neural networks (DNNs) are algorithmic computing systems that are designed after biological neural networks~\cite{pham1970neural}, e.g., the human brain.
A \emph{feed forward neural network} consists of \emph{neurons}, which are arranged in \emph{layers} that are typically densely connected to each other.
A neuron processes the linear combination of the \emph{weighted} output of \emph{all} (hence the term \emph{dense}) neurons in the previous layer.
In turn, its output is obtained by applying an \emph{activation function} to the combination of the resulting value and a \emph{bias} value.
In the end, multiple such layers form a DNN. 

The weights and biases are then adjusted to fit the data as follows. In the \emph{forward-propagation} step, part of the data (a \emph{batch}) is passed through the DNN to obtain the predictions thereof. These predictions are then evaluated according to a \emph{loss function}. To minimize the loss function, the weights and biases of the DNN are adapted using gradient descent in order to improve the quality of the predictions in the \emph{back-propagation} step. 
After the whole data set is passed through the DNN once, we say that one \emph{epoch} passed. After sufficiently many epochs, the DNN is trained. Predictions for a new object are then obtained by passing it through the DNN.

In a \emph{recurrent neural network} the current layer additionally receives its own last value as an input. TabNet~\cite{tabnet}, a neural network designed especially for tabular data, follows such a recurrent structure. In particular, each recurrent block consists of a feature transformer (a network on its own), an attentive transformer (which aggregates how much each feature has been used in the current decision step) and a mask (which ensures that the model focuses on the most important features).

\subsection{Using Location Data for DNNs\label{sec:ensemble}}
The location of a property is known to highly determine its price~\cite{Kryvobokov2007}. Hence, to make useful predictions about a property, information about the prices of the properties in its surrounding area is very useful for predictions.

Making the connections between nearby properties is usually very difficult for DNNs, which learn a direct relationship between the attributes of a property and its value.
On the other hand, CBR is all about the relationship of properties among each other, and only indirectly makes the connection to the value of a property with the help of the comparison objects.

In the case of the location-based CBR approach LBS (see Section~\ref{sec:case-based}), these relationships are captured in a single value that represents the average price in the neighborhood of a property.  Since this value can be computed easily in advance, we take it as a standard input for the considered DNNs.

We note that, instead of using the LBS values as input for the DNN, one can instead utilize the predictions made by a more sophisticated CBR approach.  Therefore, we also consider combinations of the explainable CBR+EA method and the DNNs, in order to see whether this ensemble leads to an improved performance.

\section{Empirical Comparison}
\label{sec:experiments}

We empirically evaluate the above mentioned approaches in order to answer the following questions.
\begin{enumerate}
  \item Does the explainable EA-assisted CBR approach improve the valuation performance compared to previous state-of-the-art methods? \label[hyponenum]{enum:hypothesis-ea-improvement}
  \item How does the geography of a property affect the different approaches? Is there a specific kind of area, such as cities, in which a particular approach is dominant? \label[hyponenum]{enum:hypothesis-geography}
  \item How resilient are the approaches to unclean data? \label[hyponenum]{enum:hypothesis-unclean}
  \item How does the location data provided by CBR approaches affect the performance of the DNNs? \label[hyponenum]{enum:hypothesis-ensemble}
\end{enumerate}
In the following, we first describe our experimental setup and present our results afterwards.

\subsection{Setup}
\label{sec:experiments-setup}

\paragraph{Data Set}
\begin{figure}[t]
	\centering
	\includegraphics[width=0.95\columnwidth]{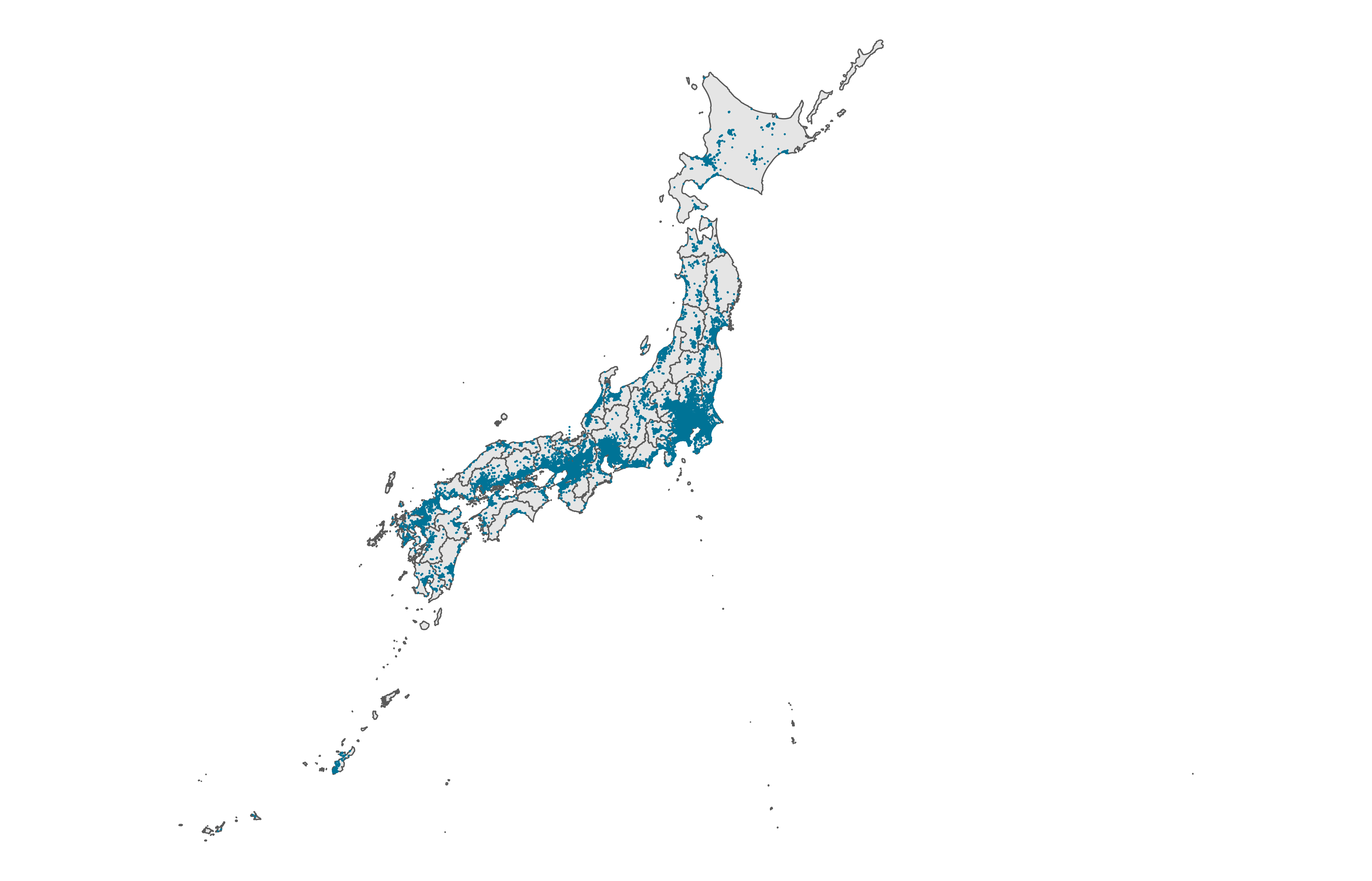}%
	\caption{Locations of properties in the considered data set.  Each blue point represents one property.}
	\label{fig:japan-sales-map}
\end{figure}
We evaluate the different approaches on a large real estate data set from Japan.
The ``LIFULL  HOME'S  Data  Set''~\cite{lifull-dataset} is available to computer science researchers worldwide upon request through the Japanese National Institute of Informatics.
It contains both rental and sales as well as residential and commercial properties.
Here, we focus on residential sales properties in the data set to test the approaches in a specific, real-world use case.

Besides standard property attributes, such as \emph{asking price}, or \emph{living area}, we use additional data included in the set, like the \emph{distance to the nearest schools and bus station}, and an \emph{urbanity score}.
Additionally, we use 12 attributes that further increase the information about a property, without reducing the size of the data set due to missing values.
The full list of attributes can be found in \Cref{tab:attributes} in~\Cref{sec:appendix}.

\label{par:data-cleaning}
Furthermore, we apply basic cleaning rules.
For properties that occur multiple times in the data set, we only consider the latest entry to prevent using the same properties in training and test set, leaving us with \num{723680} entries.
We remove outliers, i.e., data with unrealistic prices ($> \SI{300000000}{\textyen}$) or location outside of Japan.
We are then left with \num{723115} entries, whose distribution over Japan can be observed in~\Cref{fig:japan-sales-map}.

\paragraph{Experimental Setup} 

The standard approach to evaluating the performance of a method when
solving a regression task is to split the available data into two
sets: a \emph{training set} on which the method can learn and a
\emph{test set} on which it is evaluated.  Commonly, this split is
done randomly, where $80\%$ of the data are used for training, while
the remaining $20\%$ are used for testing.  Here, we consider another
split instead, which better reflects the practical applications in
real estate. In particular, the current or future price of a
property is predicted only knowing prices of past properties.  We model this use
case by splitting our data set into training and test sets at
2017-03-01, which results roughly in the common $80/20$ ratio.  All
prices posted before that date are treated as known and used for
training and all others are used as test data.
We now search for a similarity function by running the EA and train the DNNs, on the same training sets. 

The hyperparameters are mostly selected from existing best practices.
The ones chosen for the EA are described in
\Cref{tab:parameters-sim-evo}.  We picked the number of generations in
the EAs such that we did not observe or expect further improvements in
fitness beyond that point.  All CBR experiments are written in Rust
1.50.0.

Regarding the DNNs, the architecture of Kaggle Housing and Kaggle
Baseline consist of 5 and 4 densely connected layers with 200, 100,
50, 25, 1 and 128, 128, 64, 1 neurons, respectively. All layers use
\emph{relu} as activation function.  For the architecture of TabNet,
we refer the reader to the original source~\cite{tabnet}.  We chose
the number of training epochs analogous to how we determined the
number of generations for the EAs.  Kaggle Baseline and Kaggle Housing
yielded almost no improvements after $50$ epochs, while TabNet seems
to require much longer training.  Due to time constraints, we
introduced a time-out by allowing all networks to train for $200$
epochs.  For deep neural networks we employ
TensorFlow\footnote{\url{https://tensorflow.com}} 1.7.0 and Python
3.8.  A complete list of dependencies is available alongside the
source code.\footnote{A full copy of the code will be released upon
  acceptance under a license that enables free use for scientific
  purposes.}

\begin{table}[t]
  \centering
  \caption{EA Parameters. 
  The sample size defines how many predictions are made to estimate the fitness of an individual.
  The type switch probability defines how likely mutation swaps pre-selection between radius and $k$-nearest neighbors.\label{tab:parameters-sim-evo}}
  \begin{tabular}{lccc}
    \toprule
    \textbf{Parameter}      & \textbf{Value} \\
    \midrule
    Number of generations   & 200            \\
    Restart threshold       & 10             \\
    Population size         & 20             \\
    Sample size             & \num{10000}    \\
    Mutation rate           & 0.2            \\
    Type switch probability & 0.05           \\
    Offsprings per parent     & 5              \\
    \bottomrule
  \end{tabular}
\end{table}

To measure the performance, we now ask all predictors to make predictions about all properties in the test set and calculate the MAPE (\Cref{sec:error-measures}) among these.
To this end, the known similarity functions as well as the EA-produced
ones are given the training
data as valued properties \(P_{v}\).  Exemplary EA-produced similarity function parameters (i.e., the fittest individual generated in the run) can be found in \Cref{tab:actual-weights} in~\Cref{sec:appendix}.

\begin{figure*}[t]
  \begin{subfigure}{0.47\linewidth}
    \centering
    \includegraphics[width=\textwidth]{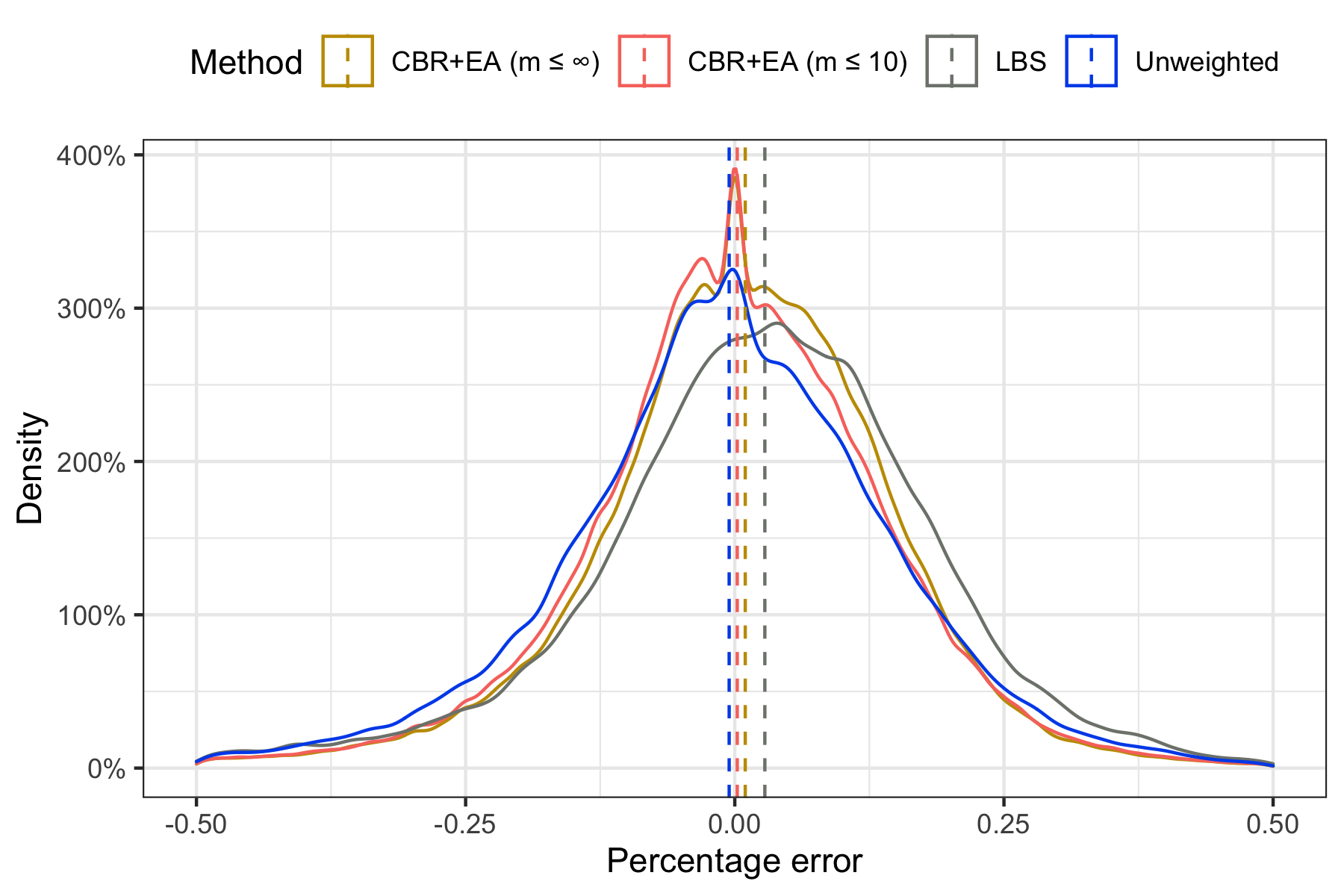}
    \caption{CBR approaches}
    \label{fig:density-cbr}
  \end{subfigure}\hspace{1em}
  \begin{subfigure}{0.47\linewidth}
    \centering
    \includegraphics[width=\textwidth]{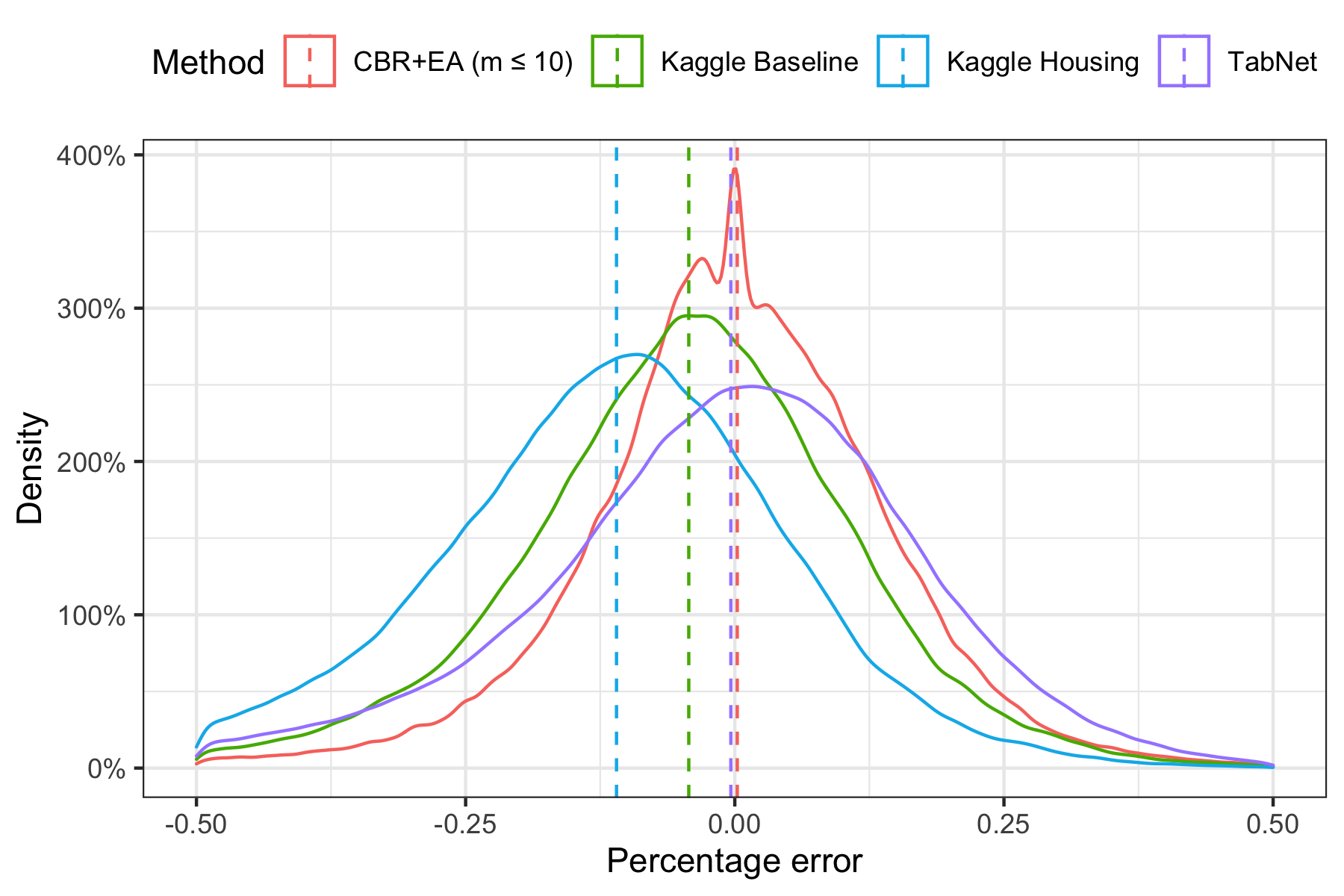}
    \caption{CBR+EA ($m \le 10$) and DNN approaches}
    \label{fig:density-dnn}
  \end{subfigure}
  \caption{Density plot of percentage errors (see Section~\ref{sec:preliminaries}).  Dashed lines show the corresponding mean percentage error (Section~\ref{sec:error-measures}).  For improved readability, the plot is cut off at 50\% percentage error (making up for 3.7 \% of predictions).}
  \label{fig:density-plots}
\end{figure*}

In total we consider seven methods in our experiments. Among them the explainable ones are
\begin{itemize}
\item Location-based similarity (LBS),
\item Unweighted similarity (Unweighted\footnote{This configuration uses the Euclidean distance, which performed best on our clean data set out of those considered by \cite{Baldominos}.}),
\item EA-learned similarity (CBR+EA).
\end{itemize}
For pre- and post-filtering (see Section~\ref{sec:case-based}), we set $r = 10$ km for LBS and Unweighted, whereas the CBR+EA method can decide between using \(k\) and~\(r\), which are constrained to \(k \le \num{2000}\) and \(r \le 50\) km, respectively.
Moreover, in preliminary experiments we determined values for $m$ that yield a good trade-off between run time and prediction quality.  We set $m = \infty$ for LBS and $m = 50$ for Unweighted, to obtain a good such trade-off.  For CBR+EA we consider two variants, one where $m \le \infty$ and one where $m \le 10$.  The latter is particularly important, as in that case the number of considered comparison objects is small enough to be evaluated by a human.

Among the unexplainable methods, we consider Kaggle Baseline, Kaggle Housing, and TabNet.

We note that all methods (except LBS and Unweighted) are inherently random, which is why each run may yield different outcomes.  To obtain meaningful statements, we perform each experiment 10 times, and report the average MAPE and the standard deviation.

\subsection{Results}
\label{sec:results}

First, we evaluate the quality of predictions generally and per prefecture before we turn our focus on analyzing the performance on unclean data, in order to determine how much the methods are affected by faulty values in the training data.
Lastly, we compare the DNNs when using different kinds of location data as input, as explained in Section~\ref{sec:ensemble}.

\subsubsection{General Predictive Performance}

\begin{figure}[b]
	\centering
	\includegraphics[width=0.95\columnwidth]{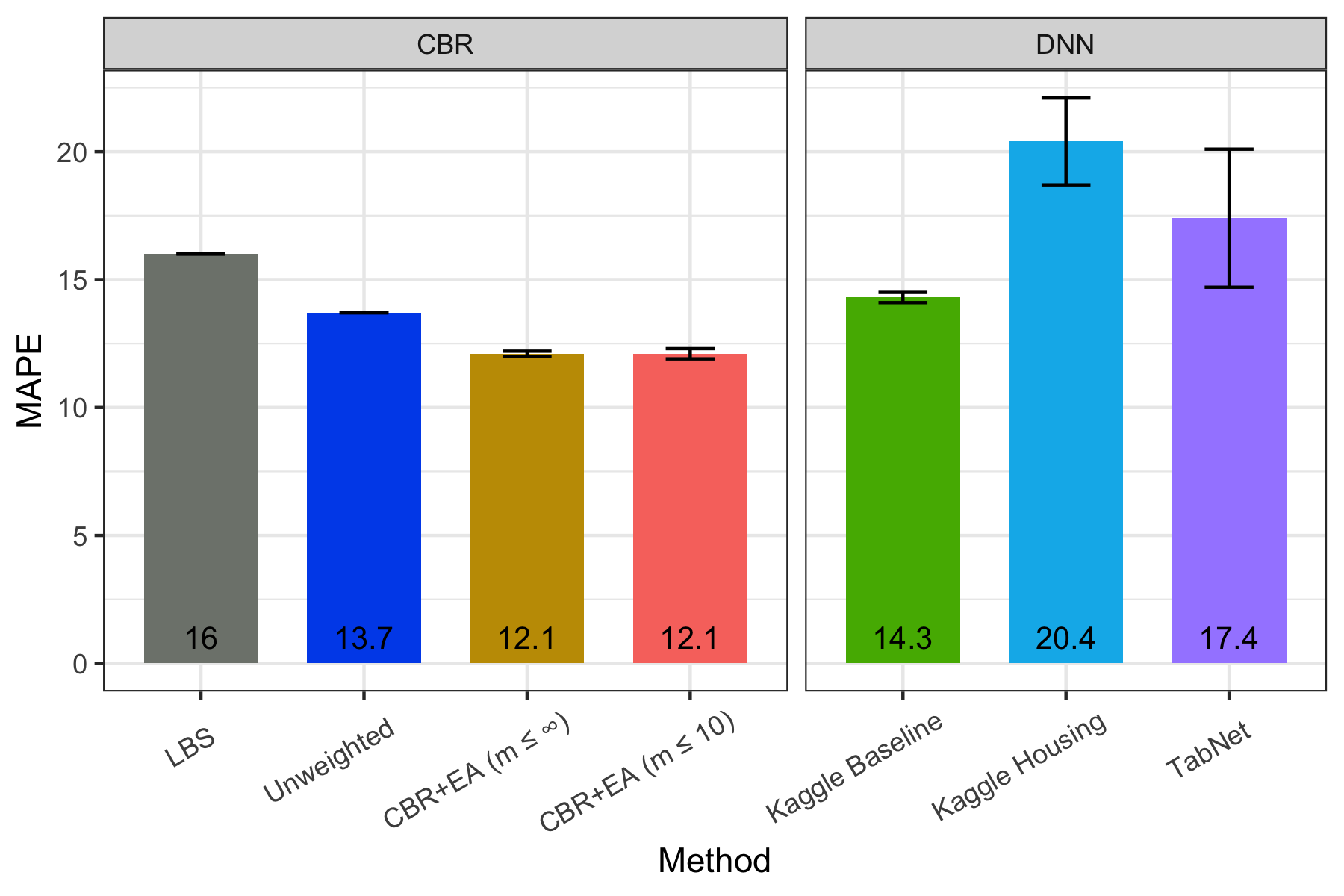}%
	\caption{Predictive performance of CBR and DNN approaches over 10
		runs. The bars denote the average MAPE in percent and the whiskers
		show the standard deviation of the MAPE over the ten runs.}
	\label{fig:general-performance}
\end{figure}

\Cref{fig:general-performance} lists performance metrics of different
DNN and CBR approaches, which we can use to answer
Question~\ref{enum:hypothesis-ea-improvement}.  As can be seen, the
LBS and Unweighted methods already yield viable results with a MAPE of
$16\%$ and $13.7\%$, deterministically.  To our surprise, the neural
networks did not provide more accuracy, only one of them getting close
to the performance of the Unweighted method, which is Kaggle Baseline
achieving a MAPE of $14.3\%$ over the 10 runs on average with a
standard deviation of $0.2\%$.  There, only one run reached a MAPE of
$13.8\%$, which is on par with the performance of the Unweighted
method.  The other two DNNs were farther off, with a MAPE of up to
$20.4\%$ on average.  Despite not being as bad, the results of TabNet
were rather unreliable with a standard deviation of $5.3\%$ over the
10 runs.

On the other hand, the EA-improved CBR approaches consistently yielded
the best results.  The CBR+EA $(m \le \infty)$ method obtained an
average MAPE of $12.1\%$ with a standard deviation of $0.1\%$.  Moreover,
the CBR+EA ($m \le 10$) variant, which uses at most~$10$ comparison
objects to compute the final prediction obtained the same results with
only a slight increase in the standard deviation~($0.2\%$). 
Thus, we obtain solid human verifiability (as less than $10$ objects are easy to comprehend) with practically no loss in performance.

Looking at the percentage errors in Figure~\ref{fig:density-plots}, we
can get a clearer picture of the prediction performances.  The LBS
method tends to overshoot the predictions, whereas the other CBR
methods, which are based on the same technique feature similar error
densities.  Notably, tuning the parameters of the similarity functions
with EAs instead of using an unweighted one, further concentrated the
density around $0$, as can be seen by the peaks in the distribution
(Figure~\ref{fig:density-cbr}).

In comparison, Figure~\ref{fig:density-dnn} reveals that the DNNs tend
to predict values that are too small, except for TabNet whose MPE is
very close to $0$.  However, its distribution is less concentrated
around $0$ than the CBR+EA method, which results in a worse MAPE.

\begin{figure*}
	\begin{subfigure}{0.47\linewidth}
		\centering
		\includegraphics[width=0.95\textwidth]{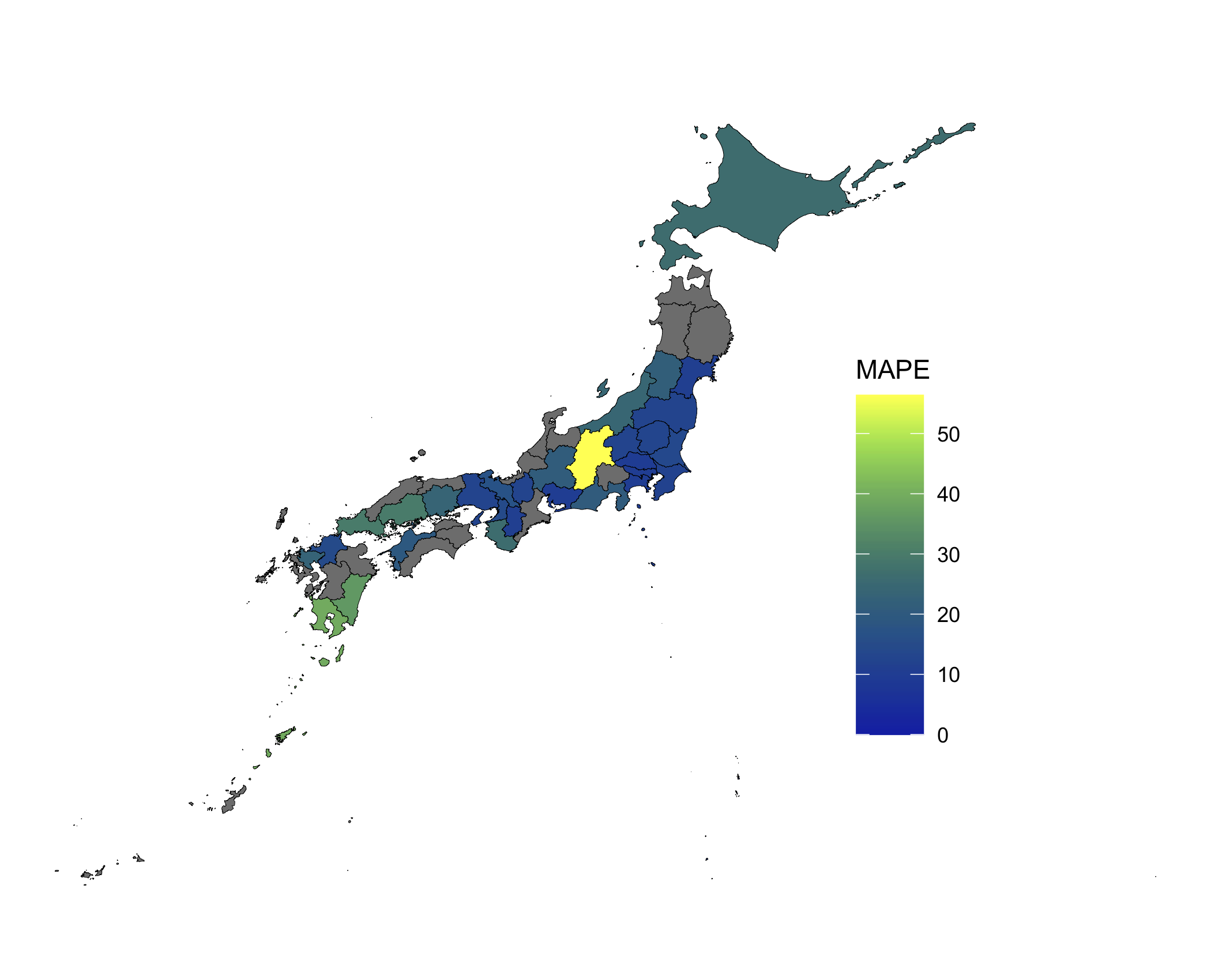}
		\caption{CBR+EA ($m \le 10$)}
		\label{fig:prefecture-mape-cbr}
	\end{subfigure}\hspace{1em}
	\begin{subfigure}{0.47\linewidth}
		\centering
		\includegraphics[width=0.95\textwidth]{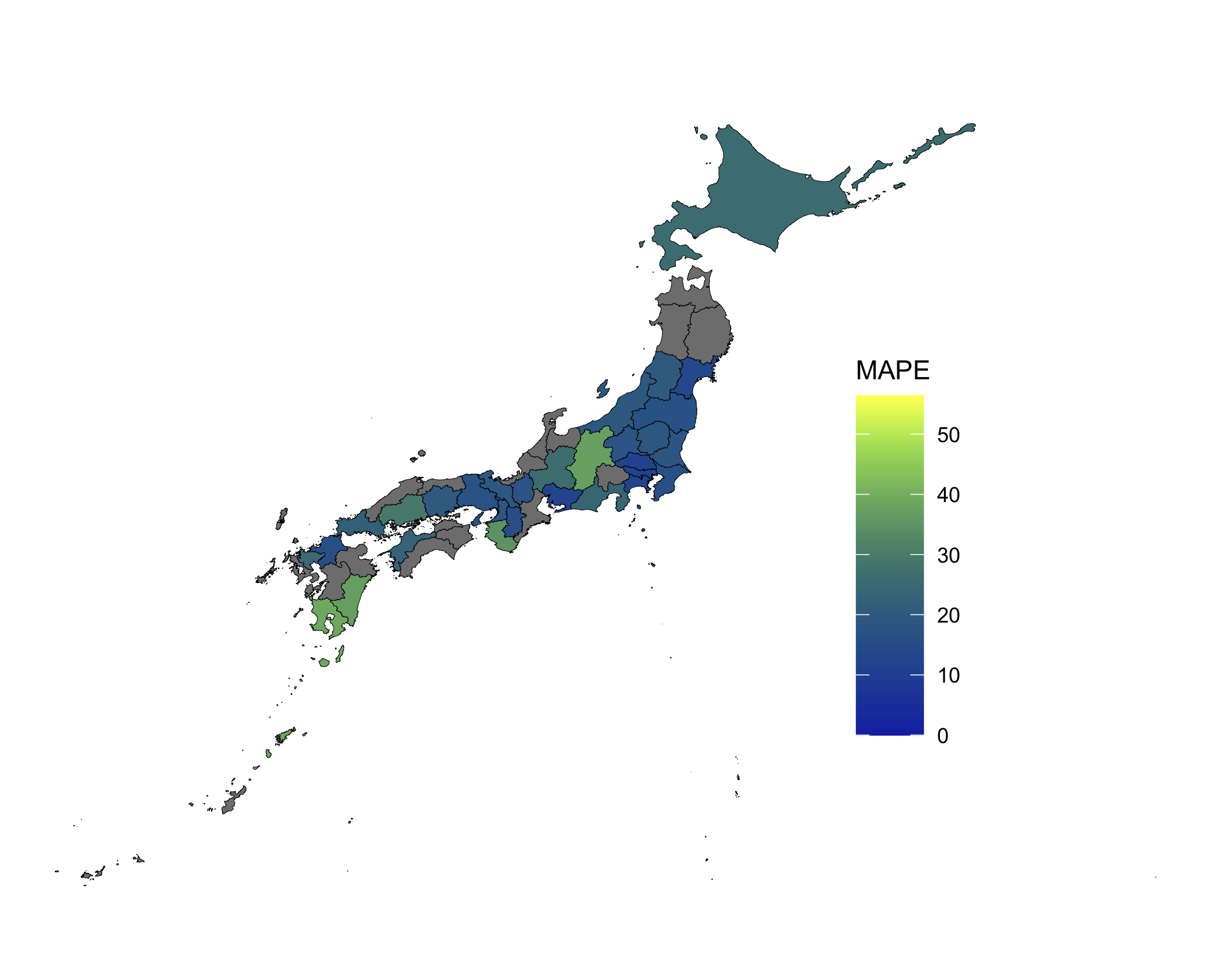}
		\caption{Kaggle Baseline}
		\label{fig:prefecture-mape-dnn}
	\end{subfigure}
	\caption{MAPE in percent per prefecture.  Prefectures with less than
		100 properties in the test set are shown in gray.}
	\label{fig:prefecture}
\end{figure*}

\subsubsection{Local Variations in Predictive Performance}
Since property prices are highly determined by location, local
variations in the available data are very likely to influence
accuracy.  To measure this, we consider the best performing CBR and
DNN methods, which are CBR+EA and Kaggle Baseline, respectively, and
compute the error measures for each of the prefectures of Japan.  To
obtain reliable results, we only consider the 30 out of 47 prefectures
in which we have at least 100 properties in the test set.  The results
are shown in Figure~\ref{fig:prefecture}.  In the following, we use
this figure to answer Question~\ref{enum:hypothesis-geography}.

While both plots look similar in many locations, there are areas where
they diverge visibly.  Most notably, for the CBR+EA predictions there
is one prefecture (Nagano, shown yellow in
Figure~\ref{fig:prefecture-mape-cbr}), which features a MAPE of
$56\%$.  On the other hand, the predictions of Kaggle Baseline in the
same prefecture yield a smaller MAPE of~$37\%$.  As can be seen in
Figure~\ref{fig:japan-sales-map}, this prefecture is rather sparsely
populated.  In fact it only contains $165$ properties, which is way
below the average number of properties per prefecture, which is ~$3749$. For CBR, this is an issue as it bases its valuations purely
on the properties in the vicinity of a property.  While this can be
advantageous, as these properties are likely especially relevant, it
also makes the approach more vulnerable to local quality differences
in the data or the number of training data close by.  On the contrary,
while DNNs take local variables into account, they are trained on the
whole data set and use that information for every prediction.
Consequently, the DNNs can be more resilient to such local variations
than CBR, which leads to a more homogeneous picture in
Figure~\ref{fig:prefecture-mape-dnn}.

If, however, the surrounding area features enough comparison objects,
CBR can utilize this local information better than DNNs, which leads
to stronger blue coloring in the densely populated prefectures
surrounding the yellow one in Figure~\ref{fig:prefecture-mape-cbr}.

\subsubsection{Unclean Data}

\begin{figure}[b]
  \centering
  \includegraphics[width=0.95\columnwidth]{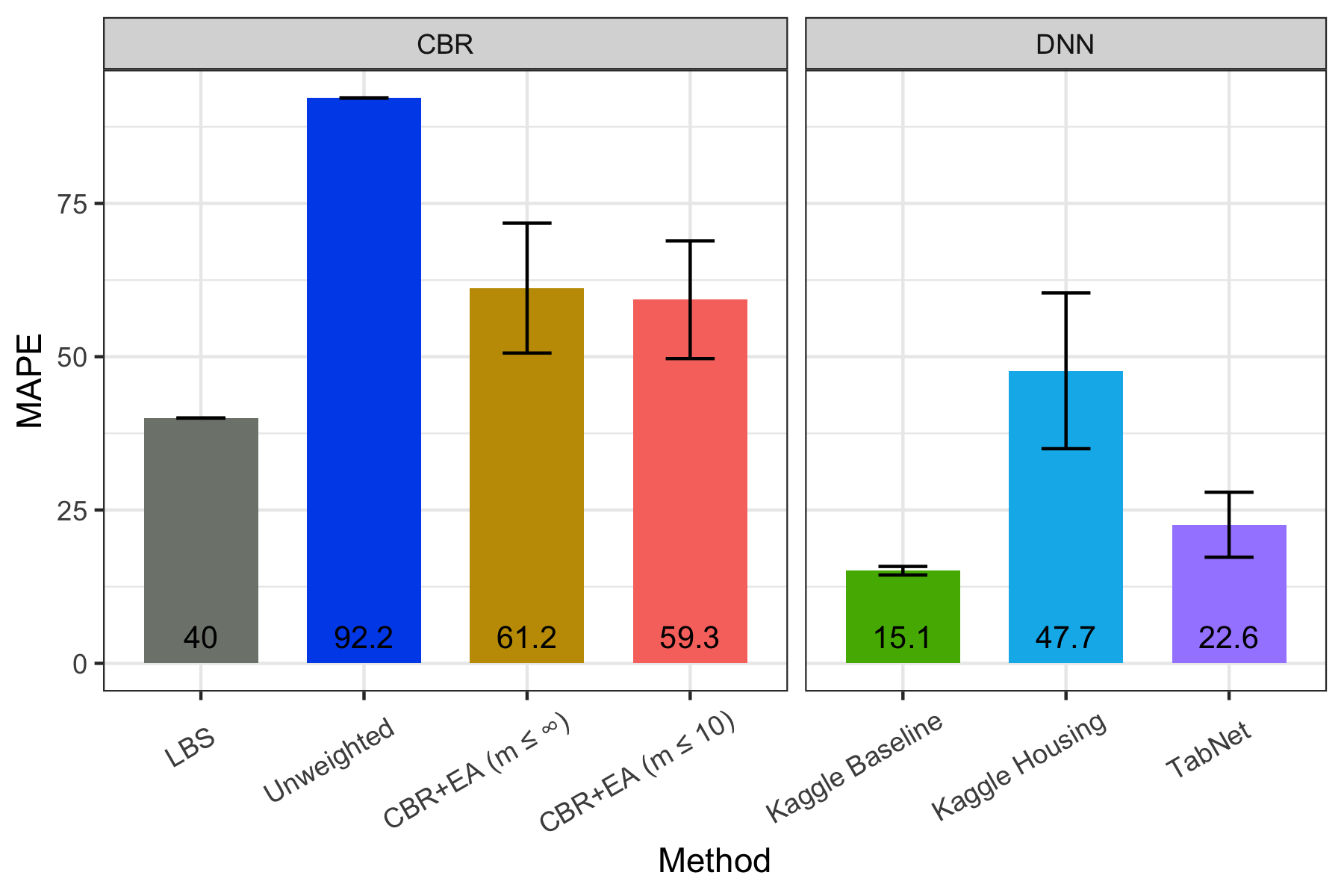}%
  \caption{Predictive performance of CBR and DNN approaches over 10
    runs on \emph{unclean data}. The bars denote the average MAPE in percent and the whiskers show the standard deviation of the MAPE over the ten runs.}
  \label{fig:performance-unclean}
\end{figure}

Since data cleaning requires lots of time and domain knowledge, the
resilience of a method to not properly cleaned data is of high
importance.  To answer Question~\ref{enum:hypothesis-unclean}, we
re-ran all our experiments without applying our cleaning steps
outlined in~\Cref{par:data-cleaning}, which adds only $565$ additional
properties as input data.

Despite this small change in the considered data, the effects are
severe, as can be observed in \Cref{fig:performance-unclean}.
For the CBR-based approaches, performance deteriorates by at least
$24\%$ (LBS), which is now the best method among the explainable ones.
In fact, for the previously well performing unweighted method the MAPE
increases by $76.5\%$ compared to the clean data, which yields a value
of $92.2\%$.  Still, improving the similarity functions with EAs
yields a strong performance increase, reducing the MAPE by more than~$30\%$ compared to the unweighted similarity function.

Similar issues with uncleaned data have been observed
before~\cite{gayer2007rule}.  Weighted average prediction is
particularly sensitive to outliers in price, due to the nature of the
average \cite{rest.88.3.443}.  A single misvaluated property can
therefore heavily influence the predictions of many similar
properties.  This effect might be mitigated by using a median instead
of an average, as explained in \cite{CELKO2015439}.

In stark contrast to the observations on the clean data, here the DNNs
can outperform all CBR approaches.  While the performance also
decreases for the DNNs, the two methods Kaggle Baseline and TabNet
only suffer a loss of $0.8\%$ and $5.2\%$ MAPE, respectively.  The
effects are stronger for Kaggle Housing, where the MAPE increased by
$27.3\%$ when using unclean data.  Still, we can conclude that the DNN
approaches tend to be more resilient to unclean data than the CBR
methods.

\subsubsection{Location Data}

\begin{figure}[t]
  \centering
  \includegraphics[width=0.95\columnwidth]{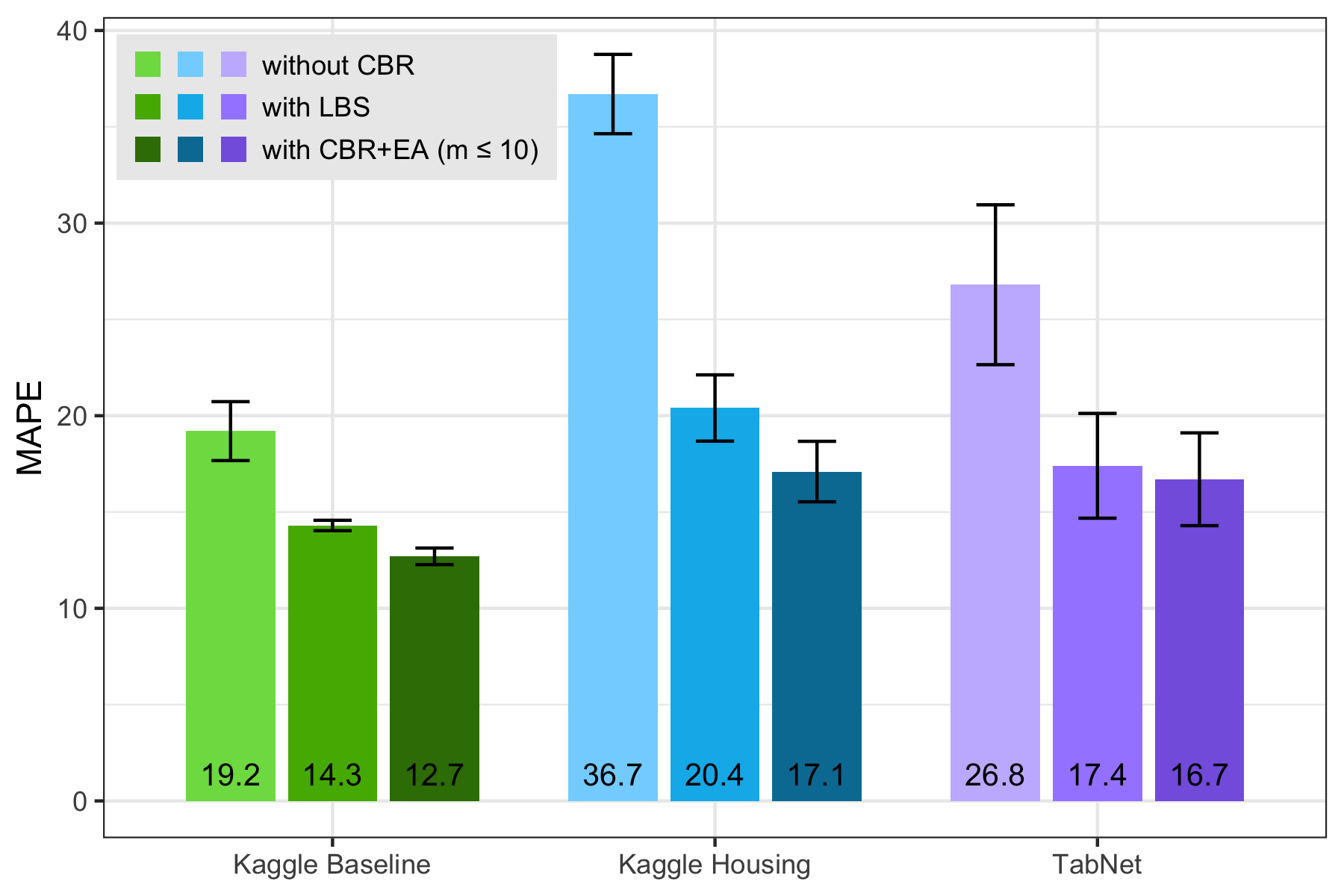}%
  \caption{Predictive performance of the DNN approaches for different inputs of location data.  Results using LBS predictions are equal to those on clean data, which are shown in \Cref{fig:general-performance}.}
  \label{fig:performance-location-data}
\end{figure}

As previously explained in Section~\ref{sec:ensemble} the location of
a property has a large impact on its value.  Therefore, we consider
the easily computable information provided by LBS to be part of the
standard input for the DNNs.  That is, we supply the DNNs with the
predictions made using LBS in all above-mentioned experiments.

To answer Question~\ref{enum:hypothesis-ensemble}, we examine how much
of an advantage this yields for the DNNs.  To this end, we performed
the experiments on clean data again, without using the LBS predictions
as input for the DNNs.  Figure~\ref{fig:performance-location-data}
shows the results.

Non-surprisingly, the performance of all DNNs decreases when the
location data is withheld.  For the best-performing DNN, Kaggle
Baseline, the increase in MAPE is smallest with $4.9\%$, yielding a
value of $19.2\%$.  As a consequence, and somewhat surprising, all
DNNs perform worse than LBS on its own, which obtained a MAPE of
$16\%$.  This highlights how reliant neural networks are on the
relationships between close-by data points captured by the LBS
information that was provided before.

As mentioned before, it is also interesting to examine whether the
performance of the DNNs increases if more sophisticated location data
is provided as input.  Therefore, we ran the same experiments but
provided the predictions of the CBR+EA ($m \le 10$) method as input
for the DNNs.  As expected, this improves the performance of the DNNs.
Kaggle Baseline now achieves a MAPE of $12.7\%$, which is better than
the standard CBR-methods.  Surprisingly, however, it does not improve
below the CBR+EA methods, whose data was provided as input.  We can
conclude that the DNN approaches are very reliant on the information
provided by the CBR predictions.

\section{Conclusion \& Future Work}
\label{sec:conclusion}

In this paper, we introduce an EA-learned CBR approach for predicting
real estate prices which, outperforms not only other CBR approaches
but also deep neural networks.  The most important factor to this
approach is the synergy between the explainable CBR method and the
optimum seeking EA.  Even though the EA itself is non-explainable, the
resulting similarity function can be interpreted and predictions made
using it can be explained with the usual CBR witness system.  Thereby,
we marry an unexplainable machine learning technique and an
explainable approach to receive both accuracy and explainability.
Through its comparison properties, our approach also enables an
interactive human-algorithm process, which allows rapid integration
into existing, human-performed real estate valuation.

Additionally, we investigate how reliant the DNNs are on location
data, which provides them with semantic information regarding
surrounding properties and greatly improves their performance.
However, some DNNs perform worse than the LBS method, even when
providing the LBS prediction as an input.  Similarly, while the
performances of the DNNs improves when allowing them to utilize the
CBR+EA predictions, they do not match the quality obtained when using
these predictions on their own.  One potential explanation is that the
data set encompassing all properties in Japan is too heterogeneous for
the DNNs to recognize and utilize local patterns.  Future work could
explore whether training different networks for the different
prefectures improves the performance in each one of them, in order to
analyze this conjecture.  Another aspect that may be investigated is
the architecture (layers, activation functions, etc.) of the
considered neural networks.  While the above considered architectures
are standard in machine learning, there are techniques to further
improve the performance of DNNs, e.g., by utilizing skip connections
or performing \emph{Neural Architecture Search (NAS)}~\cite{NAS}.  An
interesting direction would be \emph{evolutionary NAS}, which has been
used to generate more sophisticated neural networks
before~\cite{liu2020survey}.  There, EAs have previously been applied
in the context of image recognition or
classification~\cite{sun2020automatically, Fujino2017HumanSketches,
  miahi2019genetic}.  To the best of our knowledge, they have not been
applied to tabular data before.  So far, our preliminary experiments
have not shown improvements here.

We note that our combination of CBR+EA with DNNs can be seen as a
so-called \emph{ensemble method}, which refers to the combinations of
two or more machine learning algorithms that have the potential to
achieve better performance than any of them could
alone~\cite{zhang2012ensemble}.  While this was not the case in our
experiments, it would be interesting to investigate other ensembles.

While the neural networks perform worse in most of our experiments,
they were not affected as much when working with unclean data.  There,
the best performing DNN is almost not affected by the change in data
quality.  On the other hand, the performance of the CBR methods
deteriorates to a point where the predictions are unusable.  As
explained above, one explanation may be that CBR methods rely on
taking the average of determined comparison objects, which tends to be
affected heavily by faulty data.  Future work may explore how this
performance drop can be mitigated.

While considering other data like rental
prices~\cite{seya2019comparison}, or a comparison with other
techniques like hedonic price or decision tree based
models~\cite{glumac2018real,afonso2019housing}, is beyond the scope of
this paper, future work might extend our evaluation and validate our
approach using different data sets and methods.  Since we see great
potential in machine learned but explainable methods, we also look
forward to further applications in other domains, especially those
where the stakes are high.

\bibliographystyle{ACM-Reference-Format}
\bibliography{evolutionary-real-estate-valuation} 


\begin{thebibliography}{38}


\ifx \showCODEN    \undefined \def \showCODEN     #1{\unskip}     \fi
\ifx \showDOI      \undefined \def \showDOI       #1{#1}\fi
\ifx \showISBNx    \undefined \def \showISBNx     #1{\unskip}     \fi
\ifx \showISBNxiii \undefined \def \showISBNxiii  #1{\unskip}     \fi
\ifx \showISSN     \undefined \def \showISSN      #1{\unskip}     \fi
\ifx \showLCCN     \undefined \def \showLCCN      #1{\unskip}     \fi
\ifx \shownote     \undefined \def \shownote      #1{#1}          \fi
\ifx \showarticletitle \undefined \def \showarticletitle #1{#1}   \fi
\ifx \showURL      \undefined \def \showURL       {\relax}        \fi
\providecommand\bibfield[2]{#2}
\providecommand\bibinfo[2]{#2}
\providecommand\natexlab[1]{#1}
\providecommand\showeprint[2][]{arXiv:#2}

\bibitem[\protect\citeauthoryear{Afonso, Melo, Oliveira, Sousa, and
  Berton}{Afonso et~al\mbox{.}}{2019}]%
        {afonso2019housing}
\bibfield{author}{\bibinfo{person}{Bruno Afonso}, \bibinfo{person}{Luckeciano
  Melo}, \bibinfo{person}{Willian Oliveira}, \bibinfo{person}{Samuel Sousa},
  {and} \bibinfo{person}{Lilian Berton}.} \bibinfo{year}{2019}\natexlab{}.
\newblock \showarticletitle{Housing Prices Prediction with a Deep Learning and
  Random Forest Ensemble}. In \bibinfo{booktitle}{\emph{{ENIAC}}}.
  \bibinfo{pages}{389--400}.
\newblock


\bibitem[\protect\citeauthoryear{Aggarwal, Hinneburg, and Keim}{Aggarwal
  et~al\mbox{.}}{2001}]%
        {aggarwal2001surprising}
\bibfield{author}{\bibinfo{person}{Charu~C Aggarwal},
  \bibinfo{person}{Alexander Hinneburg}, {and} \bibinfo{person}{Daniel~A
  Keim}.} \bibinfo{year}{2001}\natexlab{}.
\newblock \showarticletitle{On the Surprising Behavior of Distance Metrics in
  High Dimensional Space}. In \bibinfo{booktitle}{\emph{{ICDT}}}.
  \bibinfo{pages}{420--434}.
\newblock


\bibitem[\protect\citeauthoryear{Ar{\i}k and Pfister}{Ar{\i}k and
  Pfister}{2021}]%
        {tabnet}
\bibfield{author}{\bibinfo{person}{Sercan~{\"{O}} Ar{\i}k} {and}
  \bibinfo{person}{Tomas Pfister}.} \bibinfo{year}{2021}\natexlab{}.
\newblock \showarticletitle{{TabNet}: Attentive Interpretable Tabular
  Learning}. In \bibinfo{booktitle}{\emph{{AAAI}}}.
  \bibinfo{pages}{6679--6687}.
\newblock


\bibitem[\protect\citeauthoryear{Baldominos, Blanco, Moreno, Iturrarte,
  \'{O}scar Bernárdez, and Afonso}{Baldominos et~al\mbox{.}}{2018}]%
        {Baldominos}
\bibfield{author}{\bibinfo{person}{Alejandro Baldominos},
  \bibinfo{person}{Iván Blanco}, \bibinfo{person}{Antonio~José Moreno},
  \bibinfo{person}{Rubén Iturrarte}, \bibinfo{person}{\'{O}scar Bernárdez},
  {and} \bibinfo{person}{Carlos Afonso}.} \bibinfo{year}{2018}\natexlab{}.
\newblock \showarticletitle{Identifying Real Estate Opportunities Using Machine
  Learning}.
\newblock \bibinfo{journal}{\emph{Appl. Sci.}} \bibinfo{volume}{8},
  \bibinfo{number}{11} (\bibinfo{year}{2018}).
\newblock
\urldef\tempurl%
\url{https://doi.org/10.3390/app8112321}
\showDOI{\tempurl}


\bibitem[\protect\citeauthoryear{Beckmann, Kriegel, Schneider, and
  Seeger}{Beckmann et~al\mbox{.}}{1990}]%
        {rstar}
\bibfield{author}{\bibinfo{person}{Norbert Beckmann},
  \bibinfo{person}{Hans-Peter Kriegel}, \bibinfo{person}{Ralf Schneider}, {and}
  \bibinfo{person}{Bernhard Seeger}.} \bibinfo{year}{1990}\natexlab{}.
\newblock \showarticletitle{The R*-Tree: An Efficient and Robust Access Method
  for Points and Rectangles}. In \bibinfo{booktitle}{\emph{{ICDEM}}}.
  \bibinfo{pages}{322–331}.
\newblock
\urldef\tempurl%
\url{https://doi.org/10.1145/93605.98741}
\showDOI{\tempurl}


\bibitem[\protect\citeauthoryear{Carbone and Longini}{Carbone and
  Longini}{1977}]%
        {carbone1977feedback}
\bibfield{author}{\bibinfo{person}{Robert Carbone} {and}
  \bibinfo{person}{Richard~L Longini}.} \bibinfo{year}{1977}\natexlab{}.
\newblock \showarticletitle{A Feedback Model for Automated Real Estate
  Assessment}.
\newblock \bibinfo{journal}{\emph{Manage. Sci.}} \bibinfo{volume}{24},
  \bibinfo{number}{3} (\bibinfo{year}{1977}), \bibinfo{pages}{241--248}.
\newblock


\bibitem[\protect\citeauthoryear{{\v{C}}eh, Kilibarda, Lisec, and
  Bajat}{{\v{C}}eh et~al\mbox{.}}{2018}]%
        {other_re_ml_with_mape}
\bibfield{author}{\bibinfo{person}{Marjan {\v{C}}eh}, \bibinfo{person}{Milan
  Kilibarda}, \bibinfo{person}{Anka Lisec}, {and} \bibinfo{person}{Branislav
  Bajat}.} \bibinfo{year}{2018}\natexlab{}.
\newblock \showarticletitle{Estimating the Performance of Random Forest versus
  Multiple Regression for Predicting Prices of the Apartments}.
\newblock \bibinfo{journal}{\emph{ISPRS Int. J. Geo-Inf.}} \bibinfo{volume}{7},
  \bibinfo{number}{5} (\bibinfo{year}{2018}), \bibinfo{pages}{168}.
\newblock
\urldef\tempurl%
\url{https://doi.org/10.3390/ijgi7050168}
\showDOI{\tempurl}


\bibitem[\protect\citeauthoryear{Celko}{Celko}{2015}]%
        {CELKO2015439}
\bibfield{author}{\bibinfo{person}{Joe Celko}.}
  \bibinfo{year}{2015}\natexlab{}.
\newblock \showarticletitle{Basic Aggregate Functions}.
\newblock In \bibinfo{booktitle}{\emph{{SQL} for Smarties}}.
  \bibinfo{publisher}{Elsevier}.
\newblock
\urldef\tempurl%
\url{https://doi.org/10.1016/B978-0-12-800761-7.00023-1}
\showDOI{\tempurl}


\bibitem[\protect\citeauthoryear{Dimopoulos, Tyralis, Bakas, and
  Hadjimitsis}{Dimopoulos et~al\mbox{.}}{2018}]%
        {adgeo-45-377-2018}
\bibfield{author}{\bibinfo{person}{T. Dimopoulos}, \bibinfo{person}{H.
  Tyralis}, \bibinfo{person}{N.~P Bakas}, {and} \bibinfo{person}{D.
  Hadjimitsis}.} \bibinfo{year}{2018}\natexlab{}.
\newblock \showarticletitle{Accuracy measurement of Random Forests and Linear
  Regression for mass appraisal models that estimate the prices of residential
  apartments in Nicosia, Cyprus}.
\newblock \bibinfo{journal}{\emph{Adv. Geosci.}}  \bibinfo{volume}{45}
  (\bibinfo{year}{2018}), \bibinfo{pages}{377--382}.
\newblock
\urldef\tempurl%
\url{https://doi.org/10.5194/adgeo-45-377-2018}
\showDOI{\tempurl}


\bibitem[\protect\citeauthoryear{Dixon, Halperin, and Bilokon}{Dixon
  et~al\mbox{.}}{2020}]%
        {dixon2020machine}
\bibfield{author}{\bibinfo{person}{Matthew~F Dixon}, \bibinfo{person}{Igor
  Halperin}, {and} \bibinfo{person}{Paul Bilokon}.}
  \bibinfo{year}{2020}\natexlab{}.
\newblock \bibinfo{booktitle}{\emph{Machine Learning in Finance}}.
\newblock \bibinfo{publisher}{Springer}.
\newblock


\bibitem[\protect\citeauthoryear{Ferreira}{Ferreira}{2017}]%
        {house_prices_kaggle}
\bibfield{author}{\bibinfo{person}{Diego~S Ferreira}.}
  \bibinfo{year}{2017}\natexlab{}.
\newblock \showarticletitle{Neural Network Model for House Prices (Keras)}.
\newblock \bibinfo{journal}{\emph{Kaggle}} (\bibinfo{year}{2017}).
\newblock
\newblock
\shownote{{\url{https://www.kaggle.com/diegosiebra/neural-network-model-for-house-prices-keras}}}.


\bibitem[\protect\citeauthoryear{Fujino, Mori, and Matsumoto}{Fujino
  et~al\mbox{.}}{2017}]%
        {Fujino2017HumanSketches}
\bibfield{author}{\bibinfo{person}{Saya Fujino}, \bibinfo{person}{Naoki Mori},
  {and} \bibinfo{person}{Keinosuke Matsumoto}.}
  \bibinfo{year}{2017}\natexlab{}.
\newblock \showarticletitle{Deep convolutional networks for human sketches by
  means of the evolutionary deep learning}. In
  \bibinfo{booktitle}{\emph{{IFSA-SCIS}}}. \bibinfo{pages}{1--5}.
\newblock


\bibitem[\protect\citeauthoryear{Garcia}{Garcia}{2016}]%
        {racist-machine}
\bibfield{author}{\bibinfo{person}{Megan Garcia}.}
  \bibinfo{year}{2016}\natexlab{}.
\newblock \showarticletitle{Racist in the Machine: The Disturbing Implications
  of Algorithmic Bias}.
\newblock \bibinfo{journal}{\emph{World Policy J.}} \bibinfo{volume}{33},
  \bibinfo{number}{4} (\bibinfo{year}{2016}), \bibinfo{pages}{111--117}.
\newblock
\urldef\tempurl%
\url{muse.jhu.edu/article/645268}
\showURL{%
\tempurl}


\bibitem[\protect\citeauthoryear{Gayer, Gilboa, and Lieberman}{Gayer
  et~al\mbox{.}}{2007}]%
        {gayer2007rule}
\bibfield{author}{\bibinfo{person}{Gabrielle Gayer}, \bibinfo{person}{Itzhak
  Gilboa}, {and} \bibinfo{person}{Offer Lieberman}.}
  \bibinfo{year}{2007}\natexlab{}.
\newblock \showarticletitle{Rule-Based and Case-Based Reasoning in Housing
  Prices}.
\newblock \bibinfo{journal}{\emph{B. E. J. Theor. Econ.}}  \bibinfo{volume}{7}
  (\bibinfo{year}{2007}).
\newblock


\bibitem[\protect\citeauthoryear{Gilboa, Lieberman, and Scheidler}{Gilboa
  et~al\mbox{.}}{2011}]%
        {rest.88.3.443}
\bibfield{author}{\bibinfo{person}{Itzhak Gilboa}, \bibinfo{person}{Offer
  Lieberman}, {and} \bibinfo{person}{David Scheidler}.}
  \bibinfo{year}{2011}\natexlab{}.
\newblock \showarticletitle{A similarity-based approach to prediction}.
\newblock \bibinfo{journal}{\emph{J. Econom.}} \bibinfo{volume}{162},
  \bibinfo{number}{1} (\bibinfo{year}{2011}), \bibinfo{pages}{124--131}.
\newblock
\urldef\tempurl%
\url{https://doi.org/10.1016/j.jeconom.2009.10.015}
\showDOI{\tempurl}


\bibitem[\protect\citeauthoryear{Giudice, Paola, and Forte}{Giudice
  et~al\mbox{.}}{2017}]%
        {real-estate-ga}
\bibfield{author}{\bibinfo{person}{Vincenzo~Del Giudice},
  \bibinfo{person}{Pierfrancesco~De Paola}, {and} \bibinfo{person}{Fabiana
  Forte}.} \bibinfo{year}{2017}\natexlab{}.
\newblock \showarticletitle{Using Genetic Algorithms for Real Estate
  Appraisals}.
\newblock \bibinfo{journal}{\emph{Buildings}} \bibinfo{volume}{7},
  \bibinfo{number}{2} (\bibinfo{year}{2017}).
\newblock
\urldef\tempurl%
\url{https://doi.org/10.3390/buildings7020031}
\showDOI{\tempurl}


\bibitem[\protect\citeauthoryear{Glumac and Rosiers}{Glumac and
  Rosiers}{2018}]%
        {glumac2018real}
\bibfield{author}{\bibinfo{person}{Brano Glumac} {and}
  \bibinfo{person}{Fran{\c{c}}ois~Des Rosiers}.}
  \bibinfo{year}{2018}\natexlab{}.
\newblock \showarticletitle{Real Estate and Land Property Automated Valuation
  Systems: A Taxonomy and Conceptual Model.}
\newblock \bibinfo{journal}{\emph{LISER Working Paper Series}}
  \bibinfo{volume}{9} (\bibinfo{year}{2018}).
\newblock


\bibitem[\protect\citeauthoryear{Goel, Rao, and Shroff}{Goel
  et~al\mbox{.}}{2016}]%
        {10.1257/aer.p20161028}
\bibfield{author}{\bibinfo{person}{Sharad Goel}, \bibinfo{person}{Justin~M.
  Rao}, {and} \bibinfo{person}{Ravi Shroff}.} \bibinfo{year}{2016}\natexlab{}.
\newblock \showarticletitle{Personalized Risk Assessments in the Criminal
  Justice System}.
\newblock \bibinfo{journal}{\emph{Am. Econ. Rev.}} \bibinfo{volume}{106},
  \bibinfo{number}{5} (\bibinfo{year}{2016}), \bibinfo{pages}{119--23}.
\newblock
\urldef\tempurl%
\url{https://doi.org/10.1257/aer.p20161028}
\showDOI{\tempurl}


\bibitem[\protect\citeauthoryear{Kryvobokov}{Kryvobokov}{2007}]%
        {Kryvobokov2007}
\bibfield{author}{\bibinfo{person}{Marko Kryvobokov}.}
  \bibinfo{year}{2007}\natexlab{}.
\newblock \showarticletitle{What location attributes are the most important for
  market value?}
\newblock \bibinfo{journal}{\emph{Prop. Manag.}} \bibinfo{volume}{25},
  \bibinfo{number}{3} (\bibinfo{year}{2007}), \bibinfo{pages}{257--286}.
\newblock
\urldef\tempurl%
\url{https://doi.org/10.1108/02637470710753639}
\showDOI{\tempurl}


\bibitem[\protect\citeauthoryear{Lazovich}{Lazovich}{2020}]%
        {lazovich2020does}
\bibfield{author}{\bibinfo{person}{Tomo Lazovich}.}
  \bibinfo{year}{2020}\natexlab{}.
\newblock \showarticletitle{Does Deep Learning Have Politics?}. In
  \bibinfo{booktitle}{\emph{Neurips Resistance {AI} Workshop}}.
\newblock


\bibitem[\protect\citeauthoryear{{LIFULL Co., Ltd.}}{{LIFULL Co.,
  Ltd.}}{2019}]%
        {lifull-dataset}
\bibfield{author}{\bibinfo{person}{{LIFULL Co., Ltd.}}}
  \bibinfo{year}{2019}\natexlab{}.
\newblock \bibinfo{booktitle}{\emph{{LIFULL} {HOME'S} Data Set}}.
\newblock
\newblock
\shownote{\url{https://www.nii.ac.jp/dsc/idr/en/lifull/}}.


\bibitem[\protect\citeauthoryear{Liu, Sun, Xue, Zhang, and Yen}{Liu
  et~al\mbox{.}}{2021}]%
        {liu2020survey}
\bibfield{author}{\bibinfo{person}{Yuqiao Liu}, \bibinfo{person}{Yanan Sun},
  \bibinfo{person}{Bing Xue}, \bibinfo{person}{Mengjie Zhang}, {and}
  \bibinfo{person}{Gary~G Yen}.} \bibinfo{year}{2021}\natexlab{}.
\newblock \showarticletitle{A Survey on Evolutionary Neural Architecture
  Search}.
\newblock \bibinfo{journal}{\emph{IEEE Trans. Neural Netw. Learn. Syst.}}
  \bibinfo{volume}{PP} (\bibinfo{year}{2021}).
\newblock


\bibitem[\protect\citeauthoryear{Lowrance}{Lowrance}{2015}]%
        {roy}
\bibfield{author}{\bibinfo{person}{Roy Lowrance}.}
  \bibinfo{year}{2015}\natexlab{}.
\newblock \bibinfo{title}{Predicting the Market Value of Single-Family
  Residential Real Estate}.
\newblock
\newblock


\bibitem[\protect\citeauthoryear{Melo}{Melo}{2019}]%
        {luckeciano_kaggle}
\bibfield{author}{\bibinfo{person}{Luckeciano Melo}.}
  \bibinfo{year}{2019}\natexlab{}.
\newblock \showarticletitle{{NN} With Tabular Data - Baseline}.
\newblock \bibinfo{journal}{\emph{Kaggle}} (\bibinfo{year}{2019}).
\newblock
\newblock
\shownote{{\url{https://www.kaggle.com/luckeciano/nn-with-tabular-data-baseline}}}.


\bibitem[\protect\citeauthoryear{Miahi, Mirroshandel, and Nasr}{Miahi
  et~al\mbox{.}}{2019}]%
        {miahi2019genetic}
\bibfield{author}{\bibinfo{person}{Erfan Miahi}, \bibinfo{person}{Seyed~A
  Mirroshandel}, {and} \bibinfo{person}{Alexis Nasr}.}
  \bibinfo{year}{2019}\natexlab{}.
\newblock \showarticletitle{Genetic Neural Architecture Search for automatic
  assessment of human sperm images}.
\newblock \bibinfo{journal}{\emph{CoRR}} (\bibinfo{year}{2019}).
\newblock


\bibitem[\protect\citeauthoryear{Millar}{Millar}{2011}]%
        {MillarRussellB2011MLEa}
\bibfield{author}{\bibinfo{person}{Russell~B. Millar}.}
  \bibinfo{year}{2011}\natexlab{}.
\newblock \bibinfo{booktitle}{\emph{Maximum Likelihood Estimation and
  Inference: With Examples in R, {SAS} and {ADMB}}}.
\newblock \bibinfo{publisher}{John Wiley \& Sons}.
\newblock


\bibitem[\protect\citeauthoryear{Noor}{Noor}{2020}]%
        {noor2020can}
\bibfield{author}{\bibinfo{person}{Poppy Noor}.}
  \bibinfo{year}{2020}\natexlab{}.
\newblock \showarticletitle{Can we trust {AI} not to further embed racial bias
  and prejudice?}
\newblock \bibinfo{journal}{\emph{Br. Med. J.}}  \bibinfo{volume}{368}
  (\bibinfo{year}{2020}).
\newblock


\bibitem[\protect\citeauthoryear{O'Neil}{O'Neil}{2016}]%
        {o2016weapons}
\bibfield{author}{\bibinfo{person}{Cathy O'Neil}.}
  \bibinfo{year}{2016}\natexlab{}.
\newblock \bibinfo{booktitle}{\emph{Weapons of Math Destruction: How Big Data
  Increases Inequality and Threatens Democracy}}.
\newblock \bibinfo{publisher}{Crown}.
\newblock


\bibitem[\protect\citeauthoryear{Peter, Okagbue, Obasi, and Akinola}{Peter
  et~al\mbox{.}}{2020}]%
        {reviewDNNsOnImmos}
\bibfield{author}{\bibinfo{person}{Nkolika~J Peter}, \bibinfo{person}{Hilary~I
  Okagbue}, \bibinfo{person}{Emmanuela C.~M Obasi}, {and}
  \bibinfo{person}{Adedotun~O Akinola}.} \bibinfo{year}{2020}\natexlab{}.
\newblock \showarticletitle{Review on the Application of Artificial Neural
  Networks in Real Estate Valuation}.
\newblock \bibinfo{journal}{\emph{Int. J. Adv. Trends Comput. Sci. Eng.}}
  \bibinfo{volume}{9}, \bibinfo{number}{3} (\bibinfo{year}{2020}),
  \bibinfo{pages}{2918--2925}.
\newblock
\urldef\tempurl%
\url{https://doi.org/10.30534/ijatcse/2020/66932020}
\showDOI{\tempurl}


\bibitem[\protect\citeauthoryear{Pham}{Pham}{1970}]%
        {pham1970neural}
\bibfield{author}{\bibinfo{person}{D~T Pham}.} \bibinfo{year}{1970}\natexlab{}.
\newblock \showarticletitle{Neural networks in engineering}.
\newblock \bibinfo{journal}{\emph{WIT Trans. Ecol. Environ.}}
  \bibinfo{volume}{6} (\bibinfo{year}{1970}).
\newblock


\bibitem[\protect\citeauthoryear{Rajkomar, Dean, and Kohane}{Rajkomar
  et~al\mbox{.}}{2019}]%
        {rajkomar2019machine}
\bibfield{author}{\bibinfo{person}{Alvin Rajkomar}, \bibinfo{person}{Jeffrey
  Dean}, {and} \bibinfo{person}{Isaac Kohane}.}
  \bibinfo{year}{2019}\natexlab{}.
\newblock \showarticletitle{Machine Learning in Medicine}.
\newblock \bibinfo{journal}{\emph{N. Engl. J. Med.}} \bibinfo{volume}{380},
  \bibinfo{number}{14} (\bibinfo{year}{2019}), \bibinfo{pages}{1347--1358}.
\newblock


\bibitem[\protect\citeauthoryear{Raju, Srivastava, Bisht, Sharma, and
  Kumar}{Raju et~al\mbox{.}}{2011}]%
        {raju2011development}
\bibfield{author}{\bibinfo{person}{M~Mohan Raju}, \bibinfo{person}{RK
  Srivastava}, \bibinfo{person}{Dinesh Bisht}, \bibinfo{person}{HC Sharma},
  {and} \bibinfo{person}{Anil Kumar}.} \bibinfo{year}{2011}\natexlab{}.
\newblock \showarticletitle{Development of Artificial Neural-Network-Based
  Models for the Simulation of Spring Discharge}.
\newblock \bibinfo{journal}{\emph{Adv. Artif. Intell.}}  \bibinfo{volume}{2011}
  (\bibinfo{year}{2011}), \bibinfo{pages}{1--11}.
\newblock


\bibitem[\protect\citeauthoryear{Ren, Xiao, Chang, Huang, Li, Chen, and
  Wang}{Ren et~al\mbox{.}}{2021}]%
        {NAS}
\bibfield{author}{\bibinfo{person}{Pengzhen Ren}, \bibinfo{person}{Yun Xiao},
  \bibinfo{person}{Xiaojun Chang}, \bibinfo{person}{Po{-}Yao Huang},
  \bibinfo{person}{Zhihui Li}, \bibinfo{person}{Xiaojiang Chen}, {and}
  \bibinfo{person}{Xin Wang}.} \bibinfo{year}{2021}\natexlab{}.
\newblock \showarticletitle{A Comprehensive Survey of Neural Architecture
  Search: Challenges and Solutions}.
\newblock \bibinfo{journal}{\emph{Comput. Surv.}} \bibinfo{volume}{54},
  \bibinfo{number}{4} (\bibinfo{year}{2021}), \bibinfo{pages}{1--34}.
\newblock


\bibitem[\protect\citeauthoryear{Seya and Shiroi}{Seya and Shiroi}{2021}]%
        {seya2019comparison}
\bibfield{author}{\bibinfo{person}{Hajime Seya} {and} \bibinfo{person}{Daiki
  Shiroi}.} \bibinfo{year}{2021}\natexlab{}.
\newblock \showarticletitle{A Comparison of Residential Apartment Rent Price
  Predictions Using a Large Data Set: Kriging Versus Deep Neural Network}.
\newblock \bibinfo{journal}{\emph{Geogr. Anal.}} (\bibinfo{year}{2021}).
\newblock
\urldef\tempurl%
\url{https://doi.org/10.1111/gean.12283}
\showDOI{\tempurl}


\bibitem[\protect\citeauthoryear{Sipper, Olson, and Moore}{Sipper
  et~al\mbox{.}}{2017}]%
        {som-e-17}
\bibfield{author}{\bibinfo{person}{Moshe Sipper}, \bibinfo{person}{Randal
  Olson}, {and} \bibinfo{person}{Jason Moore}.}
  \bibinfo{year}{2017}\natexlab{}.
\newblock \showarticletitle{Evolutionary computation: The next major transition
  of artificial intelligence?}
\newblock \bibinfo{journal}{\emph{BioData Min.}}  \bibinfo{volume}{10}
  (\bibinfo{year}{2017}), \bibinfo{pages}{26}.
\newblock
\urldef\tempurl%
\url{https://doi.org/10.1186/s13040-017-0147-3}
\showDOI{\tempurl}


\bibitem[\protect\citeauthoryear{Sun, Xue, Zhang, Yen, and Lv}{Sun
  et~al\mbox{.}}{2020}]%
        {sun2020automatically}
\bibfield{author}{\bibinfo{person}{Yanan Sun}, \bibinfo{person}{Bing Xue},
  \bibinfo{person}{Mengjie Zhang}, \bibinfo{person}{Gary~G Yen}, {and}
  \bibinfo{person}{Jiancheng Lv}.} \bibinfo{year}{2020}\natexlab{}.
\newblock \showarticletitle{Automatically Designing {CNN} Architectures Using
  the Genetic algorithm for Image Classification}.
\newblock \bibinfo{journal}{\emph{{IEEE} Trans. Cybern.}} \bibinfo{volume}{50},
  \bibinfo{number}{9} (\bibinfo{year}{2020}), \bibinfo{pages}{3840--3854}.
\newblock


\bibitem[\protect\citeauthoryear{Valier}{Valier}{2020}]%
        {Valier2020}
\bibfield{author}{\bibinfo{person}{Agostino Valier}.}
  \bibinfo{year}{2020}\natexlab{}.
\newblock \showarticletitle{Who performs better? {AVMs} vs hedonic models}.
\newblock \bibinfo{journal}{\emph{J. Prop. Invest. Finance}}
  \bibinfo{volume}{38}, \bibinfo{number}{3} (\bibinfo{year}{2020}),
  \bibinfo{pages}{213--225}.
\newblock
\urldef\tempurl%
\url{https://doi.org/10.1108/JPIF-12-2019-0157}
\showDOI{\tempurl}


\bibitem[\protect\citeauthoryear{Zhang and Ma}{Zhang and Ma}{2012}]%
        {zhang2012ensemble}
\bibfield{author}{\bibinfo{person}{Cha Zhang} {and} \bibinfo{person}{Yunqian
  Ma}.} \bibinfo{year}{2012}\natexlab{}.
\newblock \bibinfo{booktitle}{\emph{Ensemble Machine Learning: Methods and
  Applications}}.
\newblock \bibinfo{publisher}{Springer}.
\newblock


\end{thebibliography}

\pagebreak
\onecolumn

\appendix

\section{Appendix}
\label{sec:appendix}
\noindent We include additional tables to mainly give more context and help with the traceability of our results.
\begin{table*}[h!]
  \centering
  \caption{\label{tab:actual-weights}
    Parameters in a the weighted average prediction $\pred^{q, \mathbf{w}, \mathbf{f}, m, k, r}$ using filters as well as pre- and post-selection. These correspond to the fittest individual in one run of the respective EAs in \Cref{fig:general-performance}.
    Weights with value 0.0 or disabled filters are marked with an empty cell.
    Attributes which are left out have fixed weight of 0.0 or cannot be chosen as filters, respectively. Before applying weights, all attributes are scaled linearly to to the interval \([0,1]\).}
  \begin{subtable}[h]{0.45\textwidth}
  \begin{tabular}{lrr}
    \toprule
    \textbf{Parameter}               & \(\bm{m \le \infty}\) & \(\bm{m \le 10}\) \\
    \midrule
    Exponent \(q\)                   & 0.283                 & 0.526             \\
    \textbf{Weights $\mathbf{w}$}    &                       &                   \\
    Balcony area                     & 0.637                 & 0.629             \\
    Floor                            & 1.000                 & 0.564             \\
    Living area                      & 0.877                 & 0.069             \\
    Location (x)                     & 1.000                 & 1.000             \\
    Location (y)                     & 0.998                 & 0.345             \\
    Lot area                         & 0.762                 & 0.492             \\
    Urbanity score                   & 1.000                 & 0.995             \\
    Year of construction             &                       & 0.122             \\
    \textit{Distance to closest}     &                       &                   \\
    Public transport                      & 0.038                 & 0.036             \\
    Convenience store                & 0.470                 &                   \\
    Elementary school                &                       &                   \\
    Junior highschool                & 0.165                 & 0.009             \\
    Parking                          & 0.434                 & 0.105             \\
    \bottomrule
    \end{tabular}
    \end{subtable}
    \begin{subtable}[h]{0.45\textwidth}
    \begin{tabular}{lrr}
    \toprule
    \textbf{Filters}                 & \(\bm{m \le \infty}\) & \(\bm{m \le 10}\) \\
    \midrule
    Balcony area                     &                       &                   \\
    Date of valuation                &                       &                   \\
    Floor                            &                       &                   \\
    Object type                      & 1                     &                   \\
    Urbanity score                   &                       &                   \\
    Year of construction (in years)  & 9.601                 & 10.754            \\
    \textit{Distance to closest}     &                       &                   \\
    Bus station                      & \num{4109.871}        &                   \\
    Convenience store                &                       &                   \\
    Elementary school                &                       &                   \\
    Junior high school               &                       &                   \\
    Parking                          &                       &                   \\
    \midrule
    \textbf{Pre- and post-selection} &                       &                   \\
    \(m\)                            & 65                    & 10                \\
    \(k\)                            & \num{625}             & \num{1304}        \\
    \(r\)                            & \(\infty\)            & \(\infty\)        \\
    \bottomrule
  \end{tabular}
  \end{subtable}
\end{table*}

\begin{table*}[h!]
	\centering
	\caption{Attributes from the data set used by the predictors.}
	\label{tab:attributes}
	\begin{tabular}{ll}
		\toprule
		\textbf{Name}               & \textbf{Description}                                                          \\
		\midrule
		Asking price                & Price that is asked for the property                                          \\
        Average local price         & (For DNNs only) predicted price of the LBS approach \\
		Balcony area                & The area of a balcony (if present)                                            \\
		Offer date           & Date on which the offer was posted               \\
		Distance public transport        & Distance to the nearest bus or train station in $m$                           \\
		Distance convenience store  & Distance to the nearest convenience store in $m$                              \\
		Distance elementary school  & Distance to the nearest elementary school in $m$                              \\
		Distance junior high school & Distance to the nearest junior high school in $m$                             \\
		Distance parking            & Distance to the nearest parking spot in $m$                                   \\
		Floor                       & Floor of the property                                                         \\
		Living area                 & Living area in $m^2$                                                          \\
		Location                    & Longitude and latitude                                                        \\
		Lot area                    &  Area of land belonging to the property                                                 \\
		Object type                 & A number denoting the type of object (e.g., house, apartment)                 \\
		Property size               & The area of the property (including garden, garage, etc.)                     \\
		Urbanity score              & A score from 1 to 4 specifying the urbanization of the property's environment \\
		Year of construction        & Year in which the property has originally been build                          \\
		\bottomrule
	\end{tabular}
\end{table*}

\end{document}